%% file: main.tex
\documentclass{article}

\usepackage[final]{corl_2022} 

\usepackage[numbers]{natbib}
\usepackage{multicol}
\usepackage{caption}
\usepackage{subcaption}
\usepackage{amsfonts}
\usepackage{float}
\usepackage{bbm}
\usepackage{bm}
\usepackage{graphicx}
\usepackage{amsmath}
\usepackage[a-1b]{pdfx}

\newcommand{\rebuttal}[1]{{#1}}
\usepackage{color, soul} 

\usepackage[toc, page]{appendix}
\usepackage{makecell}
\usepackage{wrapfig}
\usepackage{hyperref}
\usepackage{tabularx}

\title{Learning to Grasp the Ungraspable with Emergent Extrinsic Dexterity}

\author{
  Wenxuan Zhou\\
  Robotics Institute\\
  Carnegie Mellon University\\
  \texttt{wenxuanz@andrew.cmu.edu} \\
  \And
  David Held\\
  Robotics Institute\\
  Carnegie Mellon University\\
  \texttt{dheld@andrew.cmu.edu} \\
}

\begin{document}
\maketitle

\vspace{-8mm}
\begin{abstract}
A \textit{simple} gripper can solve more \textit{complex} manipulation tasks if it can utilize the external environment such as pushing the object against the table or a vertical wall, known as ``Extrinsic Dexterity.'' Previous work in extrinsic dexterity usually has careful assumptions about contacts which impose restrictions on robot design, robot motions, and the variations of the physical parameters. In this work, we develop a system based on reinforcement learning (RL) to address these limitations. We study the task of “Occluded Grasping” which aims to grasp the object in configurations that are initially occluded; the robot needs to move the object into a configuration from which these grasps can be achieved. We present a system with model-free RL that successfully achieves this task using a simple gripper with extrinsic dexterity. The policy learns emergent behaviors of pushing the object against the wall to rotate and then grasp it without additional reward terms on extrinsic dexterity. We discuss important components of the system including the design of the RL problem, multi-grasp training and selection, and policy generalization with automatic curriculum. Most importantly, the policy trained in simulation is zero-shot transferred to a physical robot. It demonstrates dynamic and contact-rich motions with a simple gripper that generalizes across objects with various size, density, surface friction, and shape with a $78\%$ success rate. Videos can be found at \url{https://sites.google.com/view/grasp-ungraspable}.
\end{abstract}

\vspace{-2mm}


\input{introduction}
\input{related_work}
\input{method}

\input{experiments}
\input{conclusion}


\clearpage
\acknowledgments{This material is based upon work supported by the National Science Foundation under Grant No. IIS-1849154 and LG Electronics. We thank Daniel Seita, Thomas Weng, Tao Chen, Homanga Bharadhwaj, and Chris Paxton for the valuable feedback.}


\bibliography{references}  
\input{appendix}

\end{document}

%% file: introduction.tex
\section{Introduction}
\vspace{-2mm}

Humans have dexterous multi-fingered hands; however, similarly dexterous robot hands are expensive and fragile.  Instead, a simple hand can achieve dexterous manipulation by leveraging the environment, known as ``Extrinsic Dexterity''~\citep{dafle2014extrinsic}. For example, a simple gripper can rotate an object in-hand by pushing it against the table~\citep{chavandafle2017samplingbased}, or lifting an object by sliding it along a vertical surface~\citep{hou2020manipulation}. With the exploitation of external resources such as contact surfaces or gravity, even simple grippers can perform skillful maneuvers that are typically studied with a multi-fingered dexterous hand. 
Different from a common practice of considering the robot and an object of interest in isolation, extrinsic dexterity focuses on a holistic view of considering the interactions among the robot, the object, and the external environment.

Previous work in extrinsic dexterity has demonstrated a variety of tasks such as in-hand reorientation with a simple gripper, prehensile pushing, or shared grasping~\citep{dafle2014extrinsic, chavandafle2017samplingbased, hou2020manipulation}. However, the underlying approaches come with several limitations such as relying on hand-designed motions or primitives with limited capability of generalizing across different objects, making assumptions about contact locations and contact modes, or requiring specific gripper design~\citep{dafle2014extrinsic,chavandafle2017samplingbased, hou2020manipulation, 8462834, chavandafle2019inhand, cheng2021contact2d, cheng2021contact3d}.
Instead, we build a system with reinforcement learning (RL) to remove these limitations.
With RL, the agent can learn a closed-loop policy of how the robot should interact with the object and the environment to solve the task, taking into account both planning and control.
In addition, when trained with domain randomization, the policy can learn to be robust to different variations of physics. These properties of RL can enable extrinsic dexterity in a more general setting. 

\begin{figure}
\centering
\vspace{-5mm}
\includegraphics[width=\textwidth]{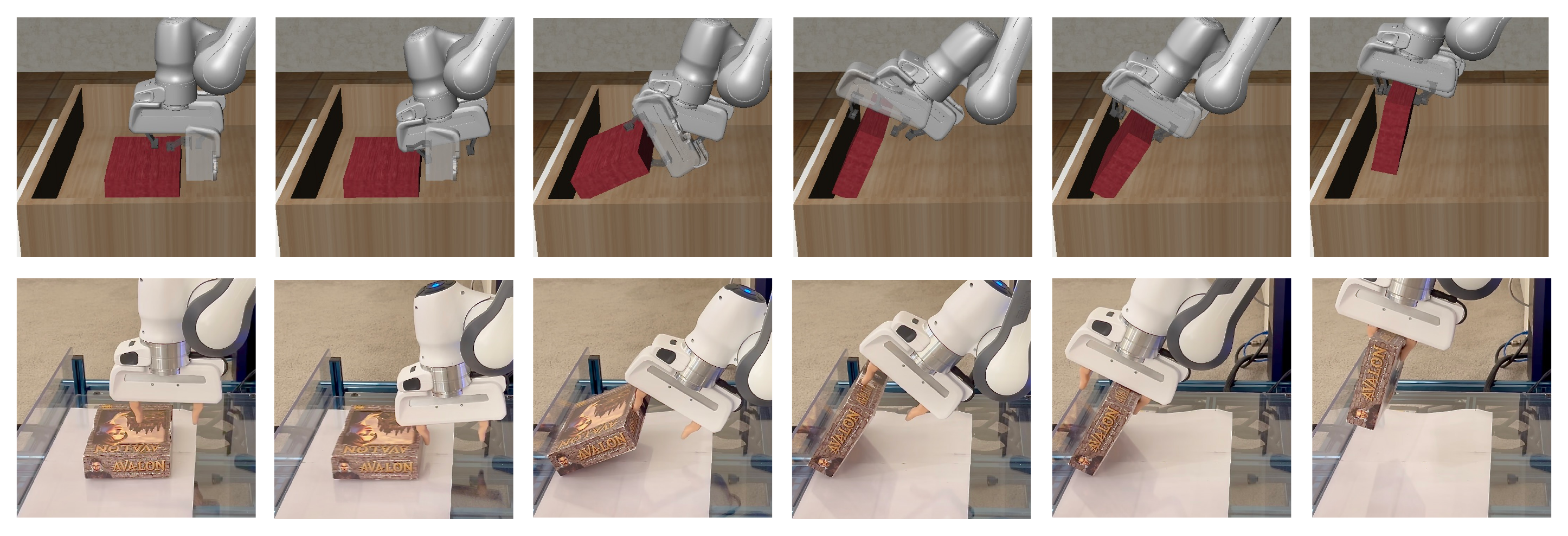}
\vspace{-6mm}
\captionof{figure}{We present a system for the ``Occluded Grasping'' task with extrinsic dexterity. The goal of this task is to reach an occluded grasp configuration (indicated by a transparent gripper). The figure shows an example of the emergent behavior of the policy and successful sim2real transfer.
\label{fig:figure1}}
\vspace{-4mm}
\end{figure}

In this work, we study ``Occluded Grasping'' as an example of a task that requires extrinsic dexterity. The goal of this task is to grasp an object in poses that are initially occluded. Consider, for example, a robot that needs to grasp a cereal box lying on its side on a table; the desired grasp is not reachable because it is partially occluded by the table (Figure~\ref{fig:figure1}). To achieve this grasp with a parallel gripper, the robot might rotate the object by pushing it against a vertical wall to expose the desired grasp and then reach it. This task is in contrast with common grasping tasks which focus on reaching an unoccluded grasp in free space with a static or near-static scene~\citep{mousavian2019graspnet, murali20206dof, wang2020manipulation}. 



The goal of this work is to build a system for the ``Occluded Grasping'' task as an example of the combination of RL and extrinsic dexterity that works on a physical robot. We investigate design choices of such a system and emphasize the simplicity of the method. With model-free RL, we design a reward function that optimizes pre-grasp and grasping motion without the separation of stages as previous work in pre-grasp~\citep{sun2020learning, chang2010planning, 8610166}. By placing the object in the bin and using a compliant low-level controller, the agent shows emergent extrinsic dexterity behavior without additional reward terms. We also incorporate a set of desired grasps with a training curriculum and a grasp selection procedure during evaluation. We improve the policy with Automatic Domain Randomization~\citep{openai2019solving} over physical parameters which robustify the contact-rich behaviors across noise and environment variations. In the experiments, we provide a comprehensive evaluation of the system in simulation to analyze the importance of each component. The policy is zero-shot transferred to the physical robot and successfully executes similar behaviors to complete the task. The policy achieves a success rate of $78\%$ and shows generalization across various out-of-distribution objects. Our main contribution is the real robot experiment results. Existing work with a simple hand has not shown such behaviors on the real robot with a similar level of complexity in contact events and generalization across objects at the same time.

%% file: related_work.tex
\vspace{-3mm}
\section{Related Work}
\vspace{-2mm}
\subsection{Extrinsic dexterity}
\vspace{-2mm}

``Extrinsic dexterity'' is a type of manipulation skill that enhances the intrinsic capability of a robot hand using external resources including external contacts, gravity, or dynamic motions of the arm~\citep{dafle2014extrinsic}. Previous work in extrinsic dexterity has demonstrated complex tasks with a simple gripper including in-hand reorientation~\citep{dafle2014extrinsic, 8462834}, prehensile pushing~\citep{chavandafle2017samplingbased, chavandafle2019inhand}, shared grasping~\citep{hou2020manipulation}, and more. 
Their methods are based on hand-crafted trajectories~\citep{dafle2014extrinsic}, task-specific motion primitives~\citep{hou2020manipulation, 8462834, 8202168}, or planning over contact modes~\citep{chavandafle2017samplingbased, chavandafle2019inhand, cheng2021contact2d, cheng2021contact3d} to simplify the problem. 
They relies on careful assumptions on contacts such as assuming a fixed number of sticking contacts between the fingertips and the object. 
In this work, we take an alternative approach to use RL to learn a closed-loop policy that considers both planning and control without limitations on contact events. The resulting policy shows more versatile contact switches beyond prior work and can be transferred to a physical robot. 

\vspace{-2mm}
\subsection{Grasping}
\vspace{-2mm}
Grasping is an important task in robot manipulation and has been studied from various aspects.

\textbf{Grasp generation:} One area of study in grasping is to generate stable grasps using analytical approaches~\citep{shimoga1996robot, nguyen1988constructing} or learning approaches~\citep{mousavian2019graspnet, murali20206dof, pinto2016supersizing, bohg2013data, murali2020object}. In our paper, we assume that the desired grasp configurations are given which may come from any grasp generation method. 

\textbf{Grasp execution:} To execute a grasp following grasp generation, a motion planner is usually used to generate a collision-free path towards the desired grasp configuration~\citep{wang2020manipulation, 5509377, 7041469,wang2021goalauxiliary}. 
All of these works aim at achieving the unoccluded grasp configurations in static or near-static scenes. Instead, our work focuses on a complementary direction of achieving occluded grasp locations by interacting with the object of interest. Another line of work in grasping uses an end-to-end pipeline without the separation of grasp generation and grasp execution~\citep{kalashnikov2018qtopt,song2020grasping}. 
However, they do not demonstrate performing the occluded grasps studied in this work.

\textbf{Pre-Grasp manipulation:} To deal with occluded grasps, prior work has studied pre-grasps as a preparatory stage of the grasping task. Typical motions for pre-grasps include rotation through planar pushing~\citep{chang2010planning}, sliding the object to the edge of the table~\citep{8610166, King-RSS-13}, or rotate the object against the wall~\citep{sun2020learning}. \citet{sun2020learning} is the most related to our work, but they use a specially designed end-effector to perform the pre-grasp motion and then use a second gripper to grasp. We demonstrate that the full grasping task can be solved with a single gripper without special requirements on the end-effectors. These previous work typically separates pre-grasp motion and grasp execution into two stages and impose restrictions on the transitions of the stages. Instead, we co-optimize pre-grasp and grasp execution throughout the episode without explicit separation of the stages. 


\vspace{-2mm}
\subsection{Reinforcement Learning for Manipulation}
\vspace{-2mm}
Previous work in RL for manipulation usually treats the object and the robot in isolation from the environment without considering extrinsic dexterity. RL has been applied to dexterous manipulation with a multi-fingered hand and shows contact-rich behaviors~\citep{openai2019solving, chen2021system, nagabandi2020deep}. In contrast, with a parallel gripper, prior work focuses on tasks with limited contacts and object motions without utilizing the environment~\citep{lee2021beyond, tobin2017domain}. This is the first work that demonstrates extrinsic dexterity with a simple parallel gripper using RL.

%% file: method.tex


\vspace{-2mm}
\section{Task Definition: Occluded Grasping}
\vspace{-2mm}
\label{task}

\begin{wrapfigure}{t}{0.5\textwidth}
    \centering
    \vspace{-3mm}
    \includegraphics[width=\linewidth]{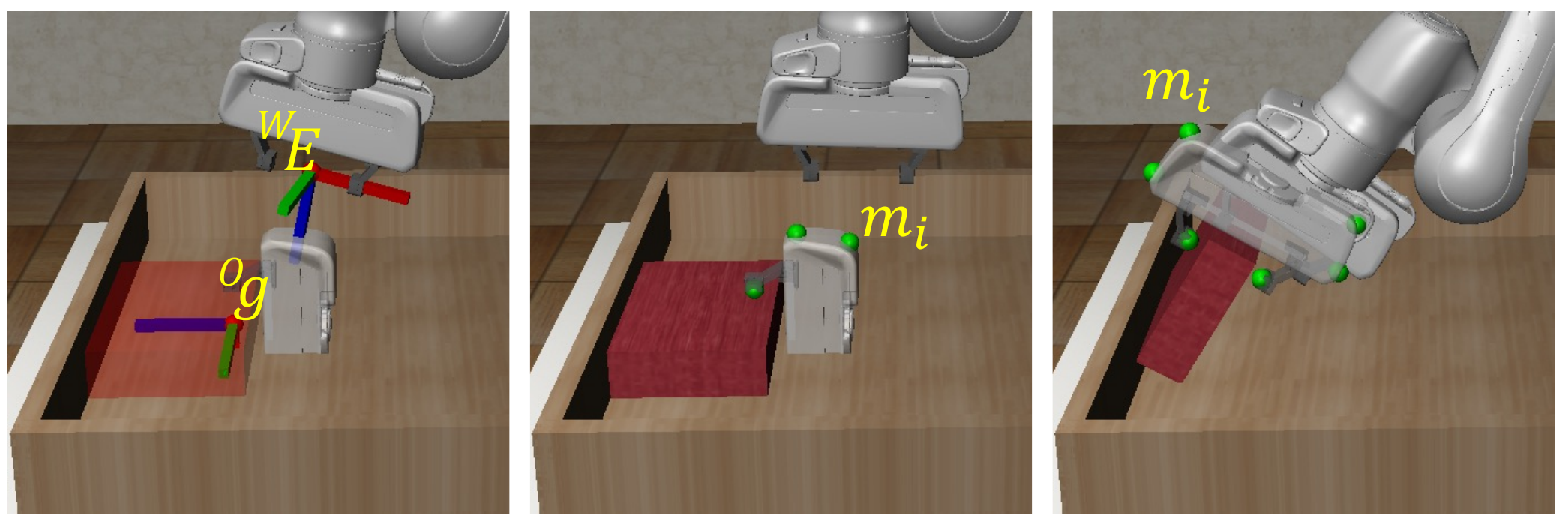}
    \caption{Notations: ${}^W E$ denotes the pose of the end-effector in the world frame ${}^W$. ${}^O g$ denotes the target grasp in the object frame ${}^O$. \rebuttal{Six marker locations} $m_i$ in green on the target grasp are used to calculate the occlusion penalty.}
    \label{fig:notations}
    \vspace{-2mm}
\end{wrapfigure}

The goal of a common grasp execution task is to move the end-effector $E$ close to a given desired grasp pose $g$. The desired grasp might come from any grasp generation method~\citep{mousavian2019graspnet, murali20206dof, murali2020object} as the input to the grasp execution task. As shown in Figure~\ref{fig:notations}, we define an \textbf{``Occluded Grasping''} task to be a subset of the grasp execution tasks where the input desired grasp $g$ is initially occluded. To clarify, the term ``occluded'' in this work is more than visual occlusion. It means the desired grasp intersects with the table and moving the gripper in free space cannot solve this task. The robot has to interact with the object to make the grasp pose reachable. The grasp ${}^O g$ is defined in the object frame $O$ and moves with the object. Formally, the grasp execution is defined to be successful if the position difference $\Delta T(g, E)$ and the orientation difference $\Delta \theta(g, E)$ are less than the pre-defined thresholds $\epsilon_T$ and $\epsilon_\theta$ respectively at the end of an episode. After successfully reaching a desired grasp pose, the gripper will be closed to complete the grasp. 
In addition, when the input to the system is a set of grasps $G=\{g_i\}_{i=1}^k$ instead of a single grasp, the agent may select any of the grasp to approach to. 



\vspace{-2mm}
\section{Learning dexterous grasping with Reinforcement Learning}
\vspace{-2mm}
We build a system that learns a closed-loop policy for the occluded grasping task defined above with model-free RL. In this section, we will discuss important design choices of the system including the design of the RL problem, the extrinsic environment, and the choice of low-level controller. Then we will discuss how to deal with a set of grasps by training with a grasp curriculum and selecting the best grasp during evaluation. We also include Automatic Domain Randomization~\citep{openai2019solving} to improving the generalization of the policy across environment variations.

\subsection{Preliminaries: Goal-conditioned Reinforcement Learning}
\label{background}

\rebuttal{
We define a Markov Decision Process (MDP) with states $s_t \in \mathcal{S}$, actions $a_t \in \mathcal{A}$, reward function $r: \mathcal{S} \times \mathcal{A} \rightarrow \mathbb{R}$, and discount factor $\gamma$. The state space, action space and the reward function for our task will be discussed in detail in the next section. The goal is to find a policy $\pi(a_t|s_t)$ that maximizes the return $R_t=\sum_{k=t}^{\infty} \gamma^{k-t}r(s_{k},a_{k})$. A Q-function is defined to be the expected return of the policy $Q^{\pi}(s,a)=\mathbb{E}_\pi[R_t|s_t,a_t]$. In goal-conditioned RL, we define a set of goals $\eta \in \mathcal{G}$ correspond to the reward function $r_{\eta}(s_t, a_t)$~\citep{schaul2015universal}. To train a policy with a set of goals, both the policy and the Q-function will now take the goal $\eta$ as input, given by $\pi(a_t|s_t, \eta)$ and $Q^{\pi}(s_t, a_t, \eta)$. In the occluded grasping task, we use the desired grasps as goals.}

\vspace{-2mm}
\subsection{RL Problem Design}
\vspace{-2mm}
\label{method:rl}

\textbf{Observation and Action Space:} The observation that is input to the policy includes a target grasp configuration in the object frame ${}^O g$, the pose of the end-effector in the world frame ${}^W E$ and the object pose in the world frame ${}^W O$. 
Note that the policy only takes one grasp ${}^O g$ as input but we will discuss how to deal with a set of grasps in Section~\ref{method:multigrasp}. For real robot experiments, we use Iterative Closest Point (ICP) for pose estimation of the object which matches a template point cloud of the object to the current point cloud~\citep{rusinkiewicz2001efficient}. 
The action space of the policy is the delta pose of the end-effector $\Delta E$ in its local frame which is passed into a low-level controller (Section~\ref{sec:OSC}).
\rebuttal{An outline of the policy execution is shown in Figure~\ref{fig:policy_duplicate}.} More details can be found in Appendix~\ref{appendix:details}.

\begin{figure}
    \centering
    \vspace{-3mm}
    \includegraphics[width=0.95\linewidth]{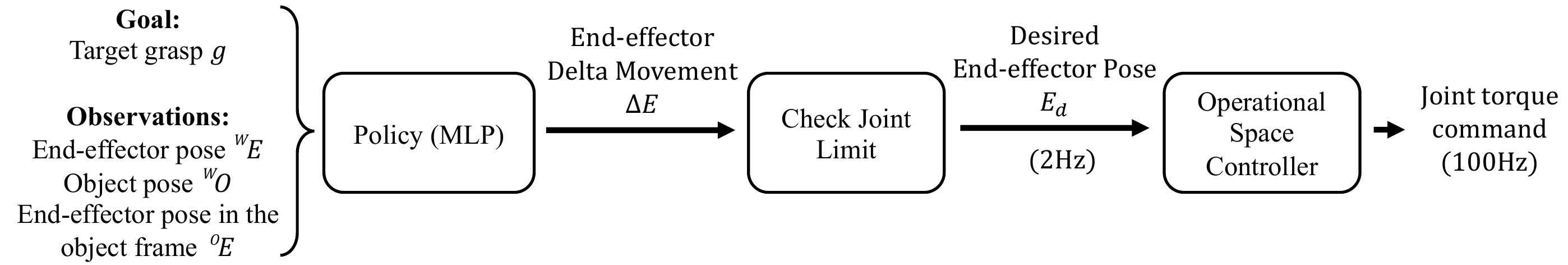}
    \caption{\rebuttal{Outline of policy execution: Given the observations, the policy outputs an end-effector delta movement (Section~\ref{method:rl}) to the low-level controller (Section~\ref{sec:OSC}).}}
    \label{fig:policy_duplicate}
    \vspace{-3mm}
\end{figure}

\textbf{Reward:}
We design the reward function to optimize the pre-grasp motion and grasp execution without separating them into two stages as in previous work~\citep{sun2020learning, 8610166, chang2010planning, King-RSS-13}:

\begin{subequations}
\begin{minipage}{.42\textwidth}
    \centering
    \begin{equation} 
    \label{eq:reward}
    r = \alpha D(g, E) + \beta \sum_{i} P(m_i)
    \end{equation}
\end{minipage}%
\begin{minipage}{.58\textwidth}
    \vspace{-4mm}
    \centering
    \begin{equation} 
    \label{eq:grasp-distance}
    D(g, E) = \alpha_1 \Delta T(g, E) + \alpha_2 \Delta \theta(g, E)
    \end{equation}
\end{minipage}
\end{subequations}
where $\alpha_1$, $\alpha_2$ and $\beta$ are the weights for the reward terms.
The first term of Equation~\ref{eq:reward},
$D(g,{}^O E)$, is the pose difference between the target grasp and the current end-effector pose, which is to optimize for reaching the grasp. This term is expanded in Equation~\ref{eq:grasp-distance} to include the translational and rotational distance, as described in Section~\ref{task}. The second term of Equation~\ref{eq:reward} is the target grasp occlusion penalty which is to penalize the agent if the target grasp configuration is in collision with the table. This corresponds to a pre-grasp objective. To measure how much the target grasp is occluded by the table, we set \rebuttal{six marker points} on the target gripper (Figure~\ref{fig:notations}) denoted as $m_i$ and compare the height of the markers with the table top. If a marker is below the table top, the height difference will be used as the penalty. Including this occlusion penalty can effectively reduce the local optima where the gripper will reach close to the target grasp (which is occluded) without trying to move the object. 
Note that we did not impose any reward terms that are explicitly related to extrinsic dexterity. In our system, the use of extrinsic dexterity is an emergent behavior of policy optimization given our objective and environmental setup.

\vspace{-2mm}
\subsection{Extrinsic Environment}
\vspace{-2mm}
\label{extrinsic_env}
To exploit the benefits of extrinsic dexterity from object-scene interaction in this task, we construct the scene as having an object in a bin, instead of leaving the object on the table as shown in Figure~\ref{fig:notations}. We also make the workspace of the robot large enough such that the robot can move the object to make contacts with the walls (during which the robot itself may also make contact with the wall). In the experiments, we will show that the policy will learn to utilize the wall to rotate the object. Without the wall, it is not able to find a strategy that can perform the task with the parallel gripper.

\vspace{-2mm}
\subsection{Choice of Low-level Controller}
\vspace{-2mm}
\label{sec:OSC}

The choice of low-level controller is important for this task due to the fact that we expect the agent to use extrinsic dexterity which involves rich contacts among the gripper, the object and the bin. We choose Operation Space Control (OSC) as the lower-level controller to execute the policy output which operates at a higher frequency (100Hz) than the RL policy (2Hz)~\citep{khatib1987unified} \rebuttal{(Figure~\ref{fig:policy_duplicate})}. Given a desired pose of the end-effector, OSC first calculates the corresponding force and torque at the end-effector to minimize the pose error according to a PD controller with gain $K_p$ and $K_d$. Then, the desired force and torque of the end-effector will be converted into desired joint torques according to the model of the robot. We choose relatively low gains so that the controller becomes compliant in the end-effector space. There are two benefits of a compliant OSC in such a contact-rich manipulation task with extrinsic dexterity. First, being compliant in end-effector space allows safe execution of the motions without smashing the gripper on the objects or the bin. Limiting the delta pose and selecting proper gains $K_p$, $K_d$ will limit the final force and torque output of the end-effector. If we use a controller that is compliant in the joint configuration space instead, we will not have direct control over the maximum force the end-effector might have on the object and the bin. Second, as shown in~\citet{martin2019variable}, using OSC as the low-level controller might speed up RL training and improve sim2real transfer for contact-rich manipulation. 

\vspace{-2mm}
\subsection{Multi-grasp Training and Grasp Selection}
\vspace{-2mm}
\label{method:multigrasp}

In this section, we consider the scenario in which a set of desired grasp configurations are given instead of just a single one. 
During training, given a set of grasps $G_{train}$, we aim at covering as many grasp configurations as possible.
As we will show in the experiments, reaching different grasps might require a significantly different behavior. Learning directly over a diverse set of goals might create difficulties for policy learning~\citep{ghosh2017divide, yu2020gradient}. We use an automatic curriculum following~\citet{openai2019solving} to gradually expand the set of grasps to be trained with. We start the training with just a single fixed grasp; after the policy has achieved a success rate larger than a threshold, it will be trained on a slightly larger set with grasps close to the initial grasp location. 

During testing, if a set of grasps $G_{test}$ is provided, we can select the best grasp within the set to improve the performance of the grasping task, following previous work in integrated grasp and motion planning~\citep{wang2020manipulation, 5509377, 7041469}. With value-based model-free RL algorithms such as Soft Actor Critic~\citep{haarnoja2018soft}, the policy is trained together with a Q-function \rebuttal{(defined in Section~\ref{background})}. We propose to select the grasp that maximizes the learned Q-function \rebuttal{for the given observation and action}: $g^* = \arg\max_{g\sim G_{test}} Q(s_t, a_t, g)$. \rebuttal{The selection can be performed for each timestep $t$, or at the beginning of an episode when $t=t_0$. We include both implementations in the experiments.} This can select the grasp that is most easily reached which depends both on the environmental configuration as well as how well the policy has learned to achieve different grasp configurations.

\vspace{-2mm}
\subsection{Improving Policy Generalization}
\vspace{-2mm}
\label{method:generalization}

To improve generalization across environment variations, we train the policy with Automatic Domain Randomization (ADR) ~\citep{openai2019solving}. Similar to the multi-grasp curriculum, the policy is first trained on a single environment with a single object, and gradually expands the range of randomization automatically according to its performance. This significantly reduces the effort of tuning the range of domain randomization. We randomize over different variations of the environment properties such as object size, density, and friction coefficient. We also randomize the parameters of the controller to improve sim2real transfer. More descriptions on the ADR procedure can be found in Appendix~\ref{appendix:adr}.

%% file: experiments.tex
\vspace{-2mm}
\section{Experiments}
\label{experiments}
\vspace{-2mm}

We build the simulation environment for this task using Robosuite~\citep{zhu2020robosuite} and the MuJoCo simulator~\citep{todorov2012mujoco} as shown in Figures~\ref{fig:figure1} and~\ref{fig:notations}. The environment contains a Franka Emika Panda robot with a parallel gripper and an object in the bin in front of it. We focus on grasping large flat objects from the side since they cannot be grasped with a top-down motion. The policy is trained in simulation with Soft Actor Critic~\citep{haarnoja2018soft}. In this section, we include the results in simulation to discuss each component of the proposed system. We then evaluate zero-shot sim2real transfer on a physical Panda robot across different objects. Implementation details can be found in Appendix~\ref{appendix:details}.

\vspace{-2mm}
\subsection{Training Curves and Ablations}
\vspace{-2mm}
\label{exp:ablations}

\begin{wrapfigure}{r}{0.6\textwidth}
\centering
\vspace{-9mm}
\begin{subfigure}{0.29\textwidth}
    \centering
    \includegraphics[width=\linewidth]{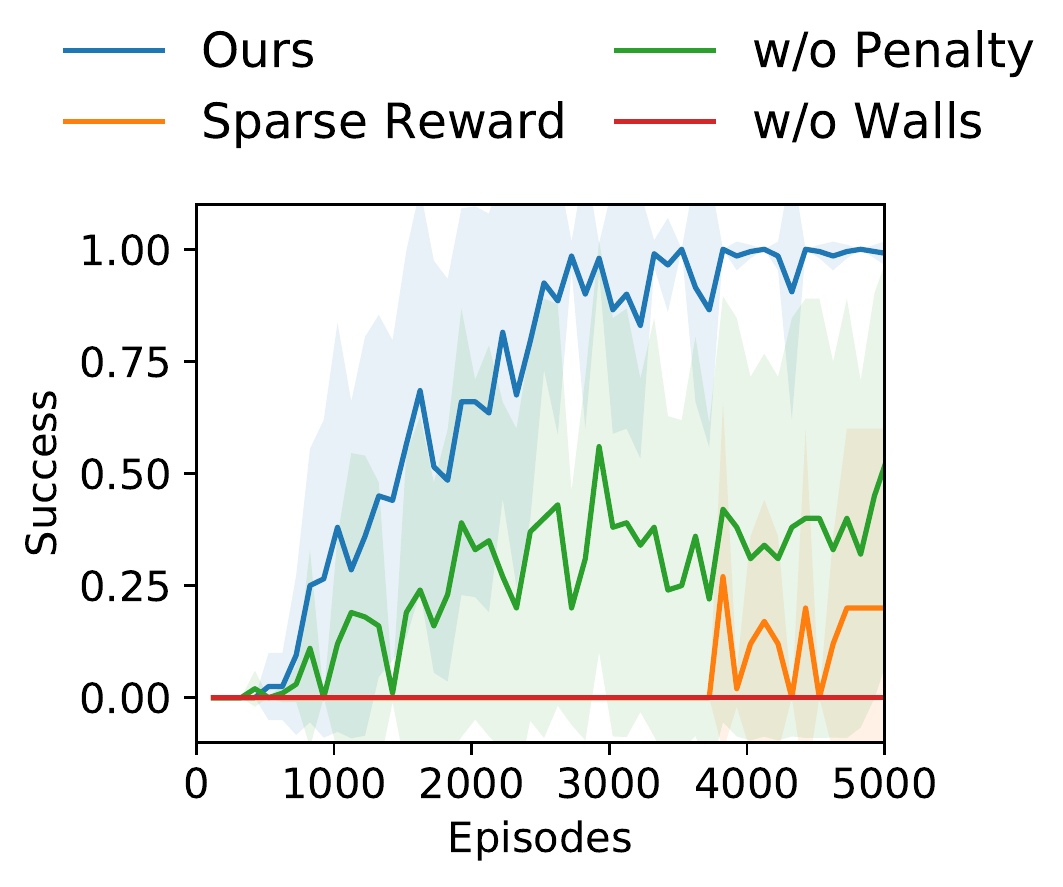}
    \vspace{-5mm}
    \label{fig:ablation}
\end{subfigure}\hspace{1mm}
\begin{subfigure}{0.29\textwidth}
    \centering
    \includegraphics[width=\linewidth]{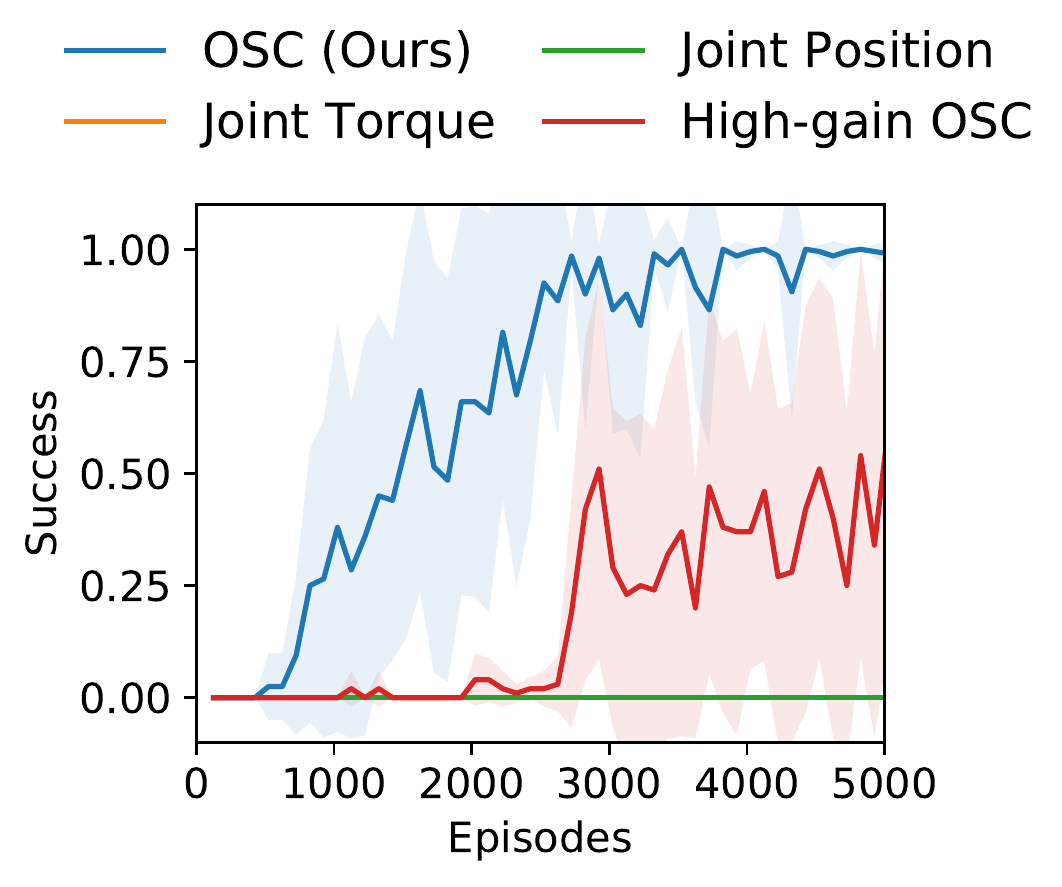}
    \vspace{-5mm}
    \label{fig:ablation_controller}
\end{subfigure}
\caption{Training curves and ablations: \textbf{(a)} ablations on the reward function and the wall. \textbf{(b)} ablations on the controller.}
\label{fig:ablations}
\vspace{-5mm}
\end{wrapfigure}

We first evaluate our method by training the policies with a single desired grasp in the default environment without ADR. Figure~\ref{fig:ablations}a shows the training curve of the proposed method and the ablations. The policy trained with the complete system can reach a success rate of $100\%$ before 4000 episodes which corresponds to 160000 environment steps. To evaluate the importance of extrinsic dexterity, we remove the walls of the bin. The resulting policies have $0\%$ success rate and push the object outside of the table. For ablations on the reward function, we remove the occlusion penalty (the second term of Equation~\ref{eq:reward}) and also try a sparse reward. Without the occlusion penalty, the policy is more likely to get stuck at a local optima (an example shown in Figure~\ref{fig:localoptima}) and thus the success rate becomes lower. 
With the alternative of a $\{-1, 0\}$ sparse reward, the policy learns much slower. 
We also experiment with different low-level controllers. Both joint torque and joint position control lead to worse performance which indicates the importance of using end-effector coordinates. With a less compliant controller by increasing the gains of OSC, the success rate becomes lower which demonstrates the importance of compliance for contact-rich tasks in addition to the safety considerations.

\vspace{-2mm}
\subsection{Emergent Behaviors}
\vspace{-2mm}


\label{exp:behavior}
Figure~\ref{fig:figure1} shows a typical strategy of a successful policy which involves multiple stages of contact switches. The gripper first moves close to the object and makes contact on the side of the object with the top finger. It then pushes the object against the wall to rotate it. During this stage, the gripper usually maintains a fixed or rolling contact with the object, but sliding also occurs. The object might have sliding or sticking contacts with the wall and the ground. After the gripper has rotated a bit further and the bottom fingertip is below the object, the gripper will let the object drop on the bottom finger. After that, the gripper will try to match the desired pose more precisely. At this point, the policy has executed the grasp successfully and it is ready to close the gripper. This type of learned contact-rich behaviors with a simple gripper has not been shown in previous work. In Section~\ref{results:realrobot}, we will further demonstrate that it can be transferred to a physical robot.

\begin{figure}
\centering
\begin{subfigure}{0.48\textwidth}
    \includegraphics[width=\linewidth]{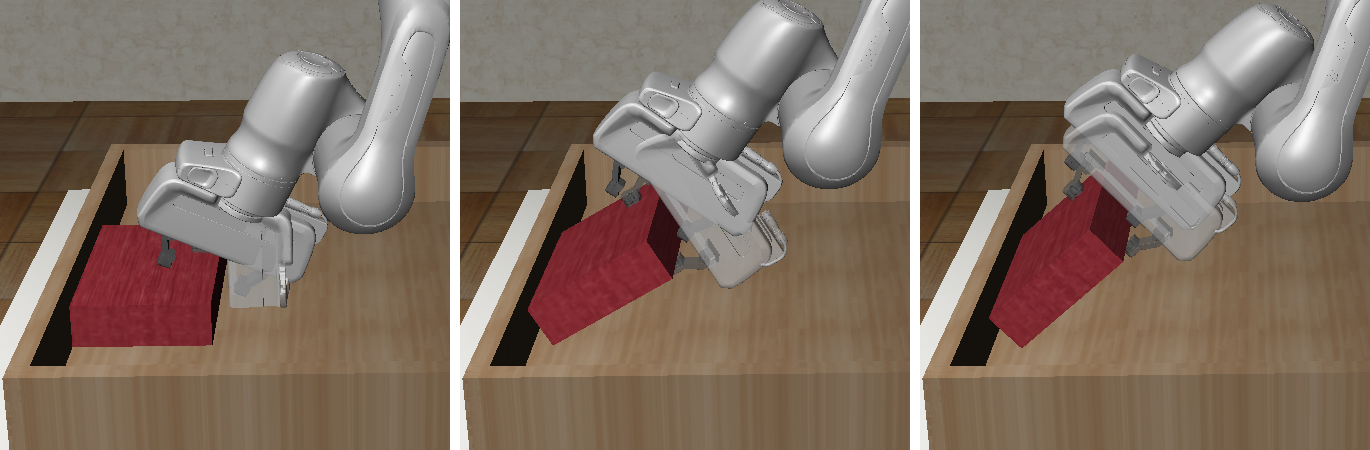}  
    \caption{\textbf{Local optima:} An example of local optima where the gripper uses the bottom finger to lift the object (instead of the top) and then fails to move the object between its two fingers to prepare for the grasp.}
    \label{fig:localoptima}
\end{subfigure}
\hfill
\begin{subfigure}{0.48\textwidth}
    \vspace{-4mm}
    \includegraphics[width=\linewidth]{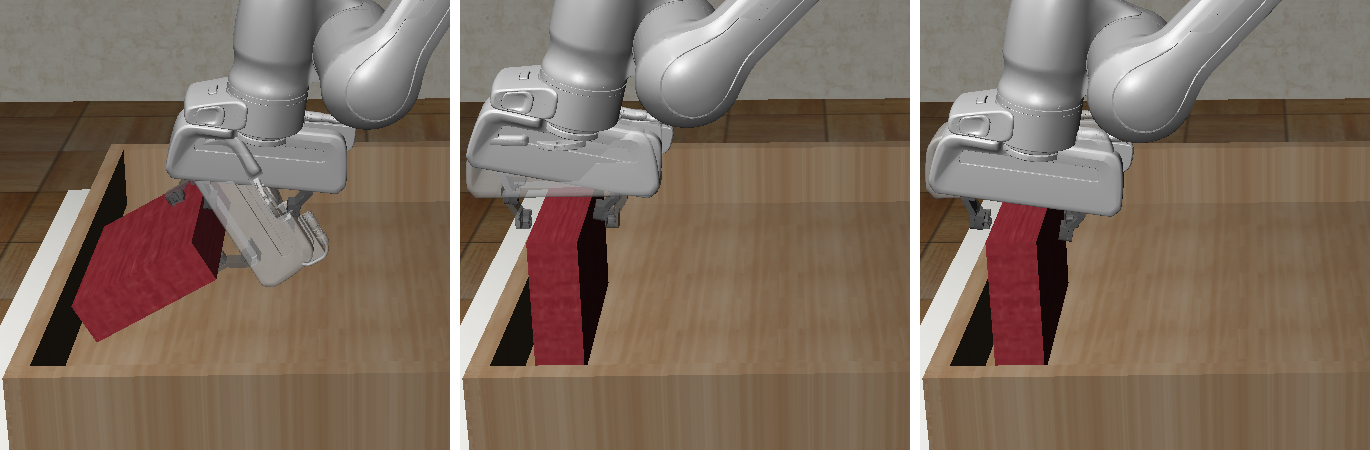}
    \caption{\textbf{Standing object:} One of the successful strategies is to flip the object until it stands on the side and then reach the grasp.}
    \label{fig:stand}
\end{subfigure}
\begin{subfigure}{0.48\textwidth}
    \includegraphics[width=\linewidth]{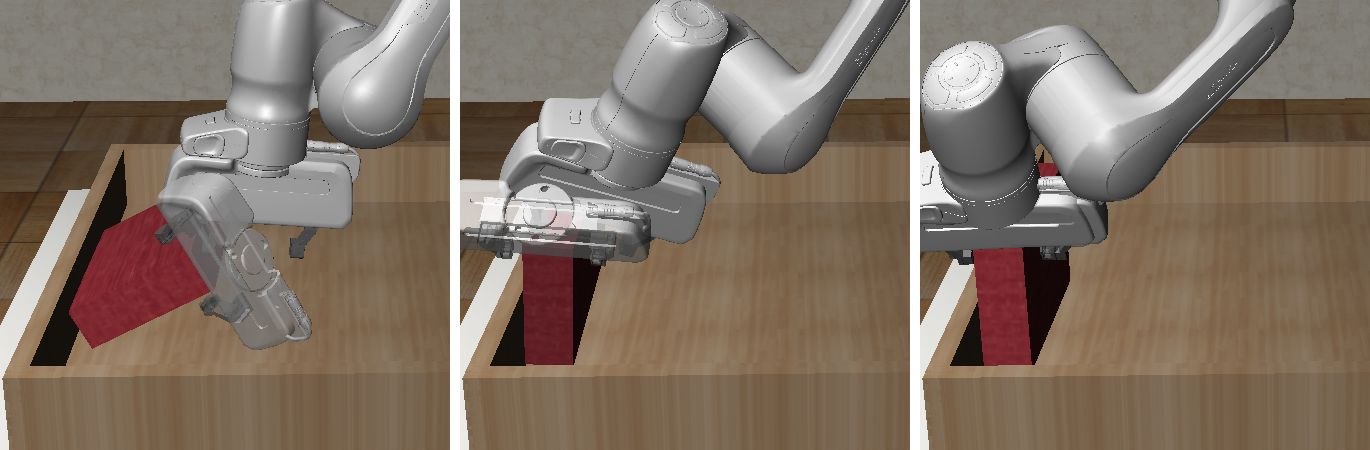}  
    \caption{\textbf{MultiGrasp-Front:} When the desired grasp is at the corner, the policy flips the object by pushing it on the side and then move close to the grasp.}
    \label{fig:multigraspfront}
\end{subfigure}
\hfill
\begin{subfigure}{0.48\textwidth}
    \includegraphics[width=\linewidth]{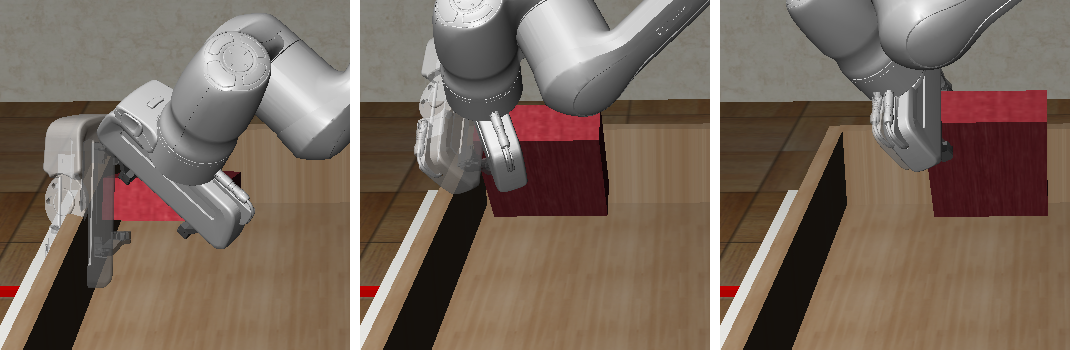}
    \caption{\textbf{MultiGrasp-Side:} When the grasp is on the side, the policy can use another side of the wall to rotate the object and reach the desired grasp.}
    \label{fig:multigraspside}
\end{subfigure}
\caption{Visualizations of the policies in different scenarios.}
\vspace{-3mm}
\label{fig:more_behaviors}
\end{figure}

One of the key decisions in this strategy is to use the top finger to rotate the object instead of the bottom finger. One might suppose an alternative approach which is to use the bottom finger to scoop the object against the wall and then directly roll the finger underneath the object to reach the grasp. However, this strategy is not physically feasible on the parallel gripper due to the limited degree of freedom of the finger. We observe that the policies that follow this strategy during exploration usually get stuck at a local optima without successfully reaching the grasp (Figure~\ref{fig:localoptima}). Another type of successful strategy is to flip the object to stand on its side and then move to the grasp (Figure~\ref{fig:stand}). This strategy relies on the fact that the object remains stable after the rotation. We will show in the real-robot experiments that for a non-box object, the object may lie on the wall to maintain stability.

\vspace{-2mm}
\subsection{Multi-grasp Experiments}
\vspace{-2mm}
\label{result:multigrasp}

\begin{figure}
  \begin{minipage}[]{.55\linewidth}
    \includegraphics[width=\linewidth]{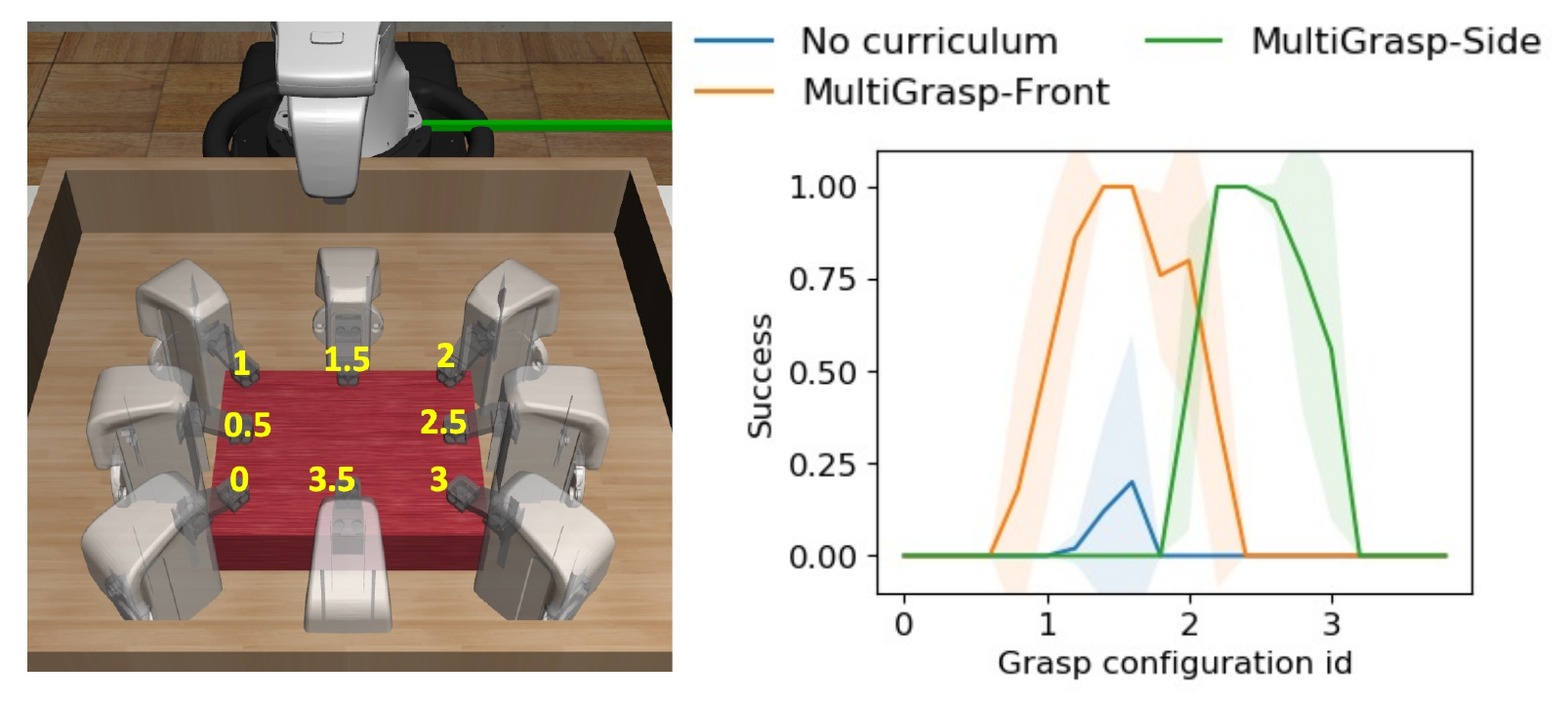}    \caption
      {%
        \textbf{Left:} Grasp configurations. \textbf{Right:} MultiGrasp Training results with and without curriculum.
        \label{fig:multigrasp}%
      }%
  \end{minipage}
  \hspace{2mm}
  \begin{minipage}[]{.4\linewidth}
  \centering
  \scalebox{0.85}{
  \begin{tabular}{lcc}
    \hline\multicolumn{1}{l}{}
    &\multicolumn{1}{c}{\bf  \makecell{MultiGrasp\\ Front}}
    &\multicolumn{1}{c}{\bf  \makecell{MultiGrasp\\ Side}}\\\hline
    ArgmaxQ	        & $1.00\pm0.00$	    & $1.00\pm0.00$\\
    ArgmaxQ-$t_0$      & $1.00\pm0.00$	    & $1.00\pm0.00$\\
    PoseDiff	    & $1.00\pm0.00$	    & $0.96\pm0.08$\\
    PoseDiff-$t_0$	    & $1.00\pm0.00$	    & $0.50\pm0.43$\\
    Uniform	        & $0.54\pm0.16$	    & $0.90\pm0.06$\\\hline
  \end{tabular}}
  \captionof{table}
      {%
        Comparison of grasp selection methods: Side grasp policies achieve better performance when using the Q-function to select the grasp.
        \label{tab:selection}%
      }
  \end{minipage}
 \vspace{-3mm}
\end{figure}

\textbf{Multi-grasp Training:} Going beyond a single desired grasp, we generate the grasp configurations around the side of the object and parameterized the grasps into a continuous grasp ID in the range of $[0,4]$ (Figure~\ref{fig:multigrasp}). We train two types of multi-grasp policies with curriculum: \textit{MultiGrasp-Front} which starts from grasp ID=$1.5$ and \textit{MultiGrasp-Side} which starts from grasp ID=$2.5$. As a baseline, we train a policy by uniformly sampling from the entire set of grasps without curriculum (\textit{No curriculum}). Figure~\ref{fig:multigrasp} shows the performance of these policies evaluated across all grasp IDs. Without curriculum, the agent has difficulties in reaching any of the grasps. With the automatic curriculum, both \textit{MultiGrasp-Front} and \textit{MultiGrasp-Side} expand from a single grasp to most of the grasps on one side of the object. Figures~\ref{fig:multigraspfront} and~\ref{fig:multigraspside} include qualitative examples of the behaviors which shows that it may require a completely different behavior for different grasps. 

\textbf{Grasp Selection:} We compare grasp selection methods with \textit{MultiGrasp-Front} and \textit{MultiGrasp-Side}. We sample 50 grasps from the training range of the policy at the beginning of each episode. The grasp selection methods will choose a grasp from this set as the input to the policy. We evaluate the following grasp selection options:
\textit{ArgmaxQ} selects the grasp that corresponds to the highest Q-value. 
\textit{PoseDiff} selects the grasp according to the closest distance to the current gripper pose according to Equation~\ref{eq:grasp-distance} (with the same weights as the reward function).
Both \textit{ArgmaxQ} and \textit{ArgmaxQ} select a grasp for each timestep. Alternatively, \textit{ArgmaxQ-$t_0$} and \textit{PoseDiff-$t_0$} only selects a grasp during the first timestep of the episode. \textit{Uniform} samples a grasp from the set uniformly. The results are summarized in Table~\ref{tab:selection}. 
For \textit{MultiGrasp-Side}, using the Q-function for grasp selection is better than the other approaches. Since the policy has a more complicated maneuver to reach the side (Figure~\ref{fig:multigraspside}), the Q-function can capture the difficulty of the goal better than pose difference. 

\subsection{Policy Generalization}
\vspace{-2mm}
\label{exp:generalization}

\begin{wrapfigure}{r}{0.48\textwidth}
\vspace{-5mm}
\includegraphics[width=\linewidth]{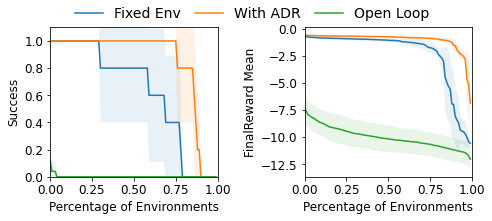}
\caption{Evaluation on the generalization of the policies by sampling 100 environments.}
\label{fig:generalization}
\vspace{-3mm}
\end{wrapfigure}

In this section, we evaluate the generalization of the policy across environment variations: open loop trajectories (\textit{Open Loop}), policies trained over a fixed environment (\textit{Fixed Env}) and policies trained with ADR (\textit{With ADR}). The open loop trajectories are obtained by rolling out the \textit{Fixed Env} policies in the default environment. 
We sample $100$ environments from the range covered by the ADR policies (Appendix~\ref{appendix:adr}) and plot the percentage of environments that are above a certain performance metric (Figure~\ref{fig:generalization}). The closed-loop policies are much better than open-loop trajectories. 
With ADR, the generalization can be improved even further. 
Sensitivity analysis on single physical parameters can be found in Appendix~\ref{appendix:results}.


\subsection{Real-robot experiments}
\label{results:realrobot}
\vspace{-2mm}

\begin{figure}
\centering
  \hspace{-8mm}
  \begin{minipage}[]{.6\linewidth}
    \centering
    \includegraphics[width=\linewidth]{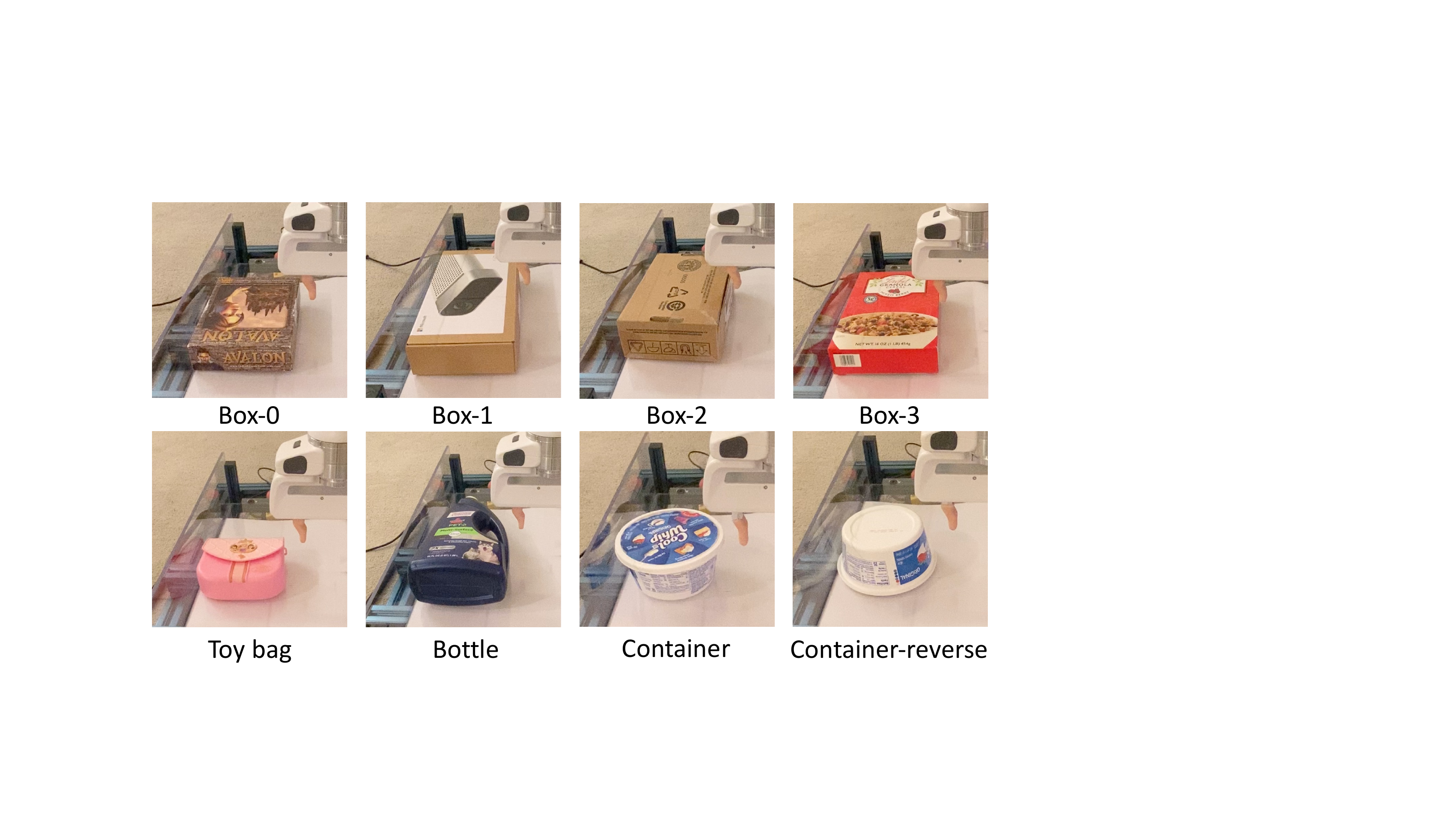}
  \end{minipage}
  \hspace{-3mm}
  \begin{minipage}[]{.4\linewidth}
  \centering
  \scalebox{0.8}{
      \label{tab:realrobot}
      \centering
      \begin{tabular}{lccc}
        \hline\multicolumn{1}{l}{\bf Object-ID}
        &\multicolumn{1}{c}{\bf \makecell{Success\\w/o ADR}}
        &\multicolumn{1}{c}{\bf \makecell{Success\\w/ ADR}}
        &\multicolumn{1}{c}{\bf \makecell{Success\\Init-pose}}\\
        
        \hline
        Box-0 (128g)	    &   9/10    &   9/10    &   7/10\\
        Box-0 (237g)        &   6/10    &  10/10    &   8/10\\
        Box-0 (345g)	    &   3/10    &   4/10    &   4/10\\
        Box-1	            &   5/10    &   8/10    &   8/10\\
        Box-2	            &   2/10    &   9/10    &   9/10\\
        Box-3	            &   0/10    &   7/10    &   9/10\\
        Toy Bag	            &   7/10    &   7/10    &   9/10\\
        Bottle	            &   0/10    &   8/10    &   1/10\\
        Container	        &   0/10    &   10/10   &   1/10\\
        Container-rev       &   0/10    &   6/10    &   0/10\\\hline
        Average	            &   33\%    &   78\%    &   56\%\\\hline
      \end{tabular}
    }
  \end{minipage}
\caption{We evaluate the policy on the real robot with various test objects. The policy trained in simulation on box-shape objects can generalize to the real robot and other shapes. With ADR, the policy achieves 45\% better success rate.}
\label{fig:objects}
\vspace{-4mm}
\end{figure}

We execute the single grasp policies on the real robot with zero-shot sim2real transfer over 10 test cases with different dimensions, densities, surface frictions, and sizes as shown in Figure~\ref{fig:objects}. \rebuttal{For non-box objects, the poses are defined with respect to the bounding boxes. The bounding boxes are obtained by running Principle Component Analysis (PCA) on the scanned object point cloud. More details of the real robot experiments can be found in Appendix~\ref{appendix:realrobot}}. Note that most of the objects are out-of-distribution.
We evaluate 10 episodes for each test case and summarize the results in Figure~\ref{fig:objects}. The success is measured by being able to close the gripper and lift the object at the end of the episode. We first compare the policies with and without Automatic Domain Randomization, denoted as \textbf{w ADR} and \textbf{w/o ADR} respectively. Quantitatively, the policy with ADR achieves a success rate of $78\%$ while the policy without ADR achieves $33\%$. 
Interestingly, the policy with ADR achieves 24/30 successes over the bottle, the Cool Whip container, and the container with a reversed initial pose. This demonstrates that although the policy is only trained with boxes in simulation, it can also generalize to other shapes to some extent when we represent the object with its bounding box. However, when the object with an out-of-distribution shape has a very different transition dynamics, the policy could fail. Qualitatively, both policies being evaluated exhibit similar strategies as discussed in Section~\ref{exp:behavior}. In fact, a single policy network may execute either the dropping strategy (Figure~\ref{fig:figure1}) or the standing strategy (Figure~\ref{fig:stand}) depending on the current state. 
We also include additional results when the initial object location is not close to the wall, denoted as \textbf{Init-pose} in Figure~\ref{fig:objects}. We finetune the \textbf{w/ ADR} policy to expand further over the range of initial object locations. The success rate remains similar for most objects, but this setting becomes more challenging for non-box shapes.
Videos of the full real robot evaluation including failure cases and recovery behaviors can be found on the website~\footnote{\url{https://sites.google.com/view/grasp-ungraspable}}. These real robot results are valuable to the field of manipulation because it is beyond what has been shown with a simple hand considering the combined complexity of contact events, object motion and object generalization.


%% file: conclusion.tex
\vspace{-2mm}
\section{Limitations}
\vspace{-2mm}

One limitation of this work is that the policy is trained with box-shape objects. Although it may generalize to other shapes to some extent as shown in the experiments, the policy might be improved by including other shapes during training. In addition, the pose of the object alone may not be sufficient to generalize to novel objects; using a better representation of the shape such as a point cloud or key-points could improve generalization across shapes. However, these changes would also increase the training complexity. \rebuttal{Another limitation is that we assume a reasonably accurate robot and gripper model, in terms of geometries, kinematic and dynamic parameters. It would be interesting to explore how to extend the method to transfer across robots and grippers. }

\vspace{-2mm}
\section{Conclusion and Takeaways}
\vspace{-2mm}

In this work, we study the ``Occluded Grasping'' task where the robot with a parallel gripper aims to reach a grasp configuration using extrinsic dexterity. We present a system that learns a closed-loop policy for this task with reinforcement learning. In the experiments, we demonstrate the importance of each component of the system.
We also show that the policies can be executed on the real robot and generalize to various objects. One potential extension of our work is to train the policy with a wide variety of object shapes which may require image-based or point cloud-based policies. Also, the pipeline can potentially be extended to other extrinsic dexterity tasks.

Despite the simplicity of the proposed method, we would like to emphasize the following takeaways from this work: First, we provide a concrete example that a simple gripper can do much more than pick-and-place while being cheaper and easier to maintain than a dexterous hand, following previous work in extrinsic dexterity. We envision more future work in this direction in manipulation. Second, RL can be a good option to generate policies with emergent extrinsic dexterity, and sim2real transfer works reasonably well with our proposed system. Our work takes a step towards deploying contact-rich policies with a simple gripper in the real world.

%% file: appendix.tex
\newpage
\begin{appendices}

\section{Additional Results}
\label{appendix:results}
\subsection{Sensitivity analysis on physical parameters}

\begin{figure}[H]
\centering
\vspace{-2mm}
\includegraphics[width=0.6\linewidth]{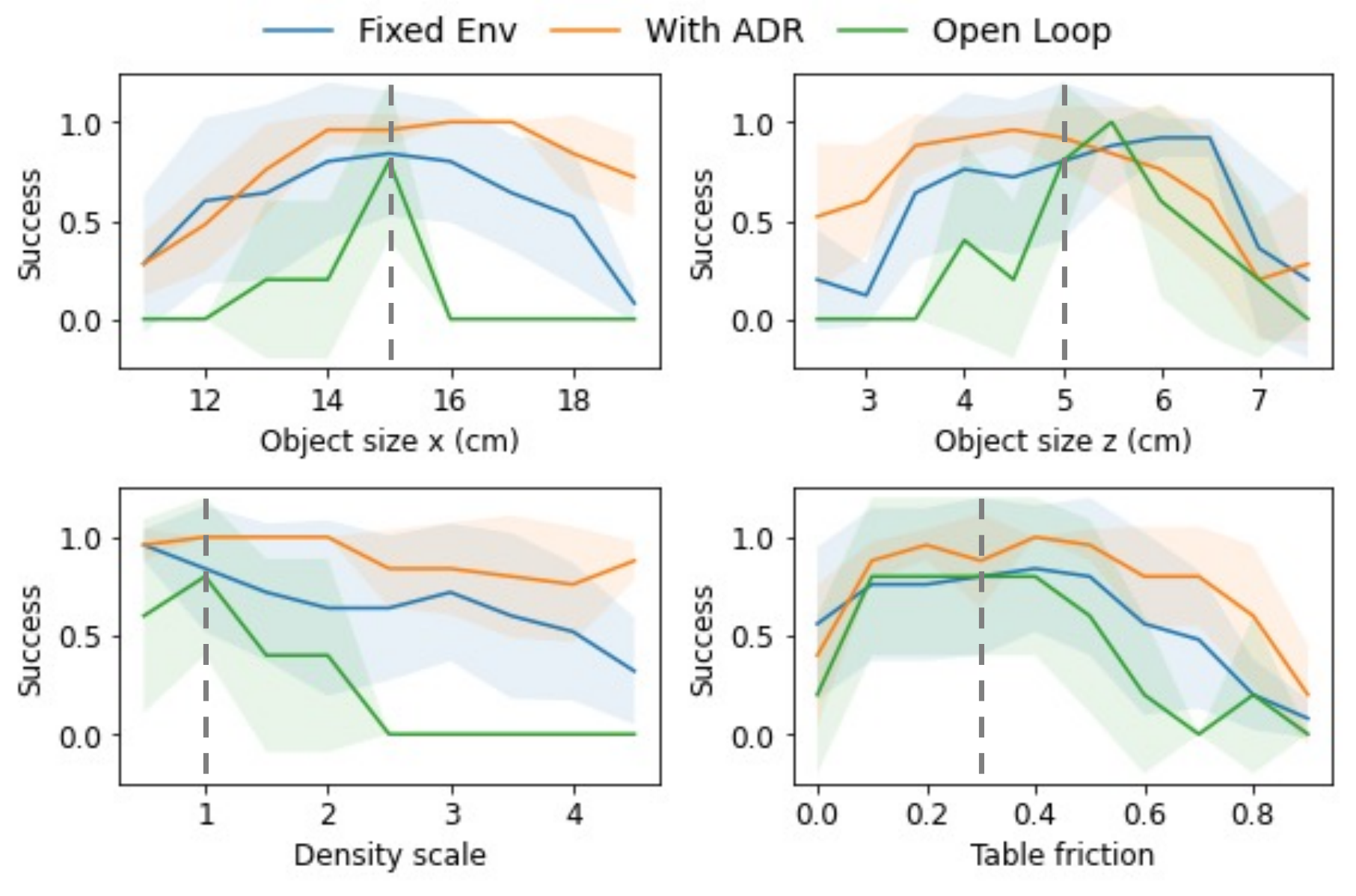}
\caption{We evaluate the generalization of policies by changing one parameter at a time. The dashed lines indicate the default values of these parameters in the fixed environment.}
\vspace{-2mm}
\label{fig:eval_params}
\end{figure}

In addition to the evaluation on policy generalization in Section~\ref{exp:generalization}, we modify the important physical parameters one at a time to understand the sensitivity of the policy performance to these parameters (Figure~\ref{fig:eval_params}). We compare the same baselines as Section~\ref{exp:generalization}: policies trained over a fixed environment (\textit{Fixed Env}), policies trained with ADR (\textit{With ADR}) and open-loop trajectories generated by rolling out the fixed env policy in the default environment (\textit{Open Loop}). The ranges of parameters are chosen to create a performance drop for all the baselines as a stress test. Similar to what we observe in Section~\ref{exp:generalization}, the policy can cover a wider range of physical parameters with closed-loop execution and with ADR.

\subsection{Sensitivity analysis on object pose estimation noise}

\begin{figure}[H]
\centering
\vspace{-2mm}
\includegraphics[width=0.6\linewidth]{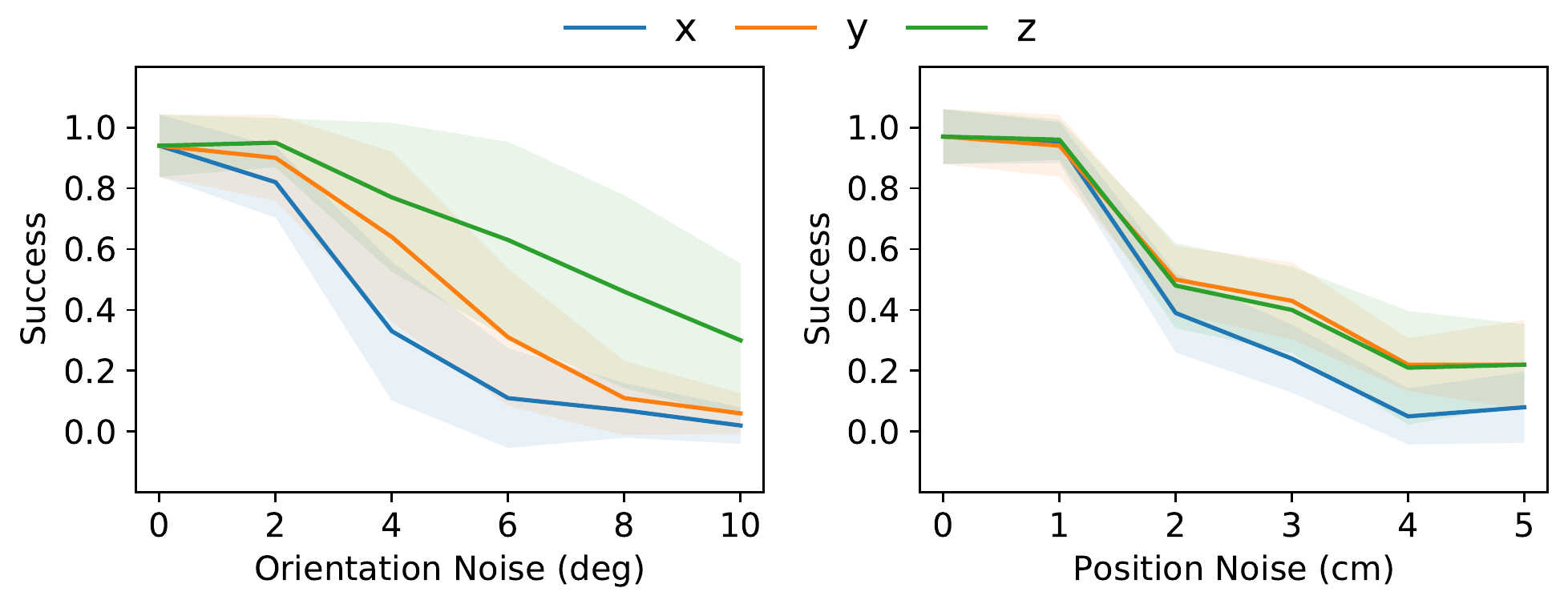}
\caption{We evaluate the sensitivity of the ADR policies on object pose estimation noise.}
\vspace{-2mm}
\label{fig:noise}
\end{figure}

\rebuttal{The proposed system takes the 6D object pose as policy input. In the real world, object pose estimation might be noisy. In this section, we evaluate the policies trained with ADR with different levels of pose estimation noise for each dimension of the 6D object pose (Figure~\ref{fig:noise}). During evaluation, for each timestep across the episode, we sample a scalar noise from a Gaussian distribution $\mathcal{N}(\mu=0, \sigma=x)$ and add it to one dimension of the object pose. The standard deviations $\sigma=x$ are shown as the x-axis in the plots. The shaded area indicates the standard deviation of the success rates across seeds.} 

\subsection{Reward term weights}

\rebuttal{The reward function shown in Equation~\ref{eq:reward} is composed of three terms with weights $\alpha_1$, $\alpha_2$ and $\beta$. $\alpha_1$ and $\alpha_2$ weight the translation and rotation error between the target grasp and the current end-effector.$\beta$ weights the target grasp occlusion penalty which is to penalize the agent if the target grasp configuration is in collision with the table. We use $\alpha_1=50$, $\alpha_2=2$, $\beta=200$ in all the experiments. In this section, we train the policies with different weight values to see how much reward tuning is required to achieve reasonable performance for the occluded grasping task. Figure~\ref{fig:weights} below shows that the policy is not too sensitive in most of the case we tested except that a higher $\alpha_1=70$ leads to a $50\%$ drop in performance.}

\begin{figure}[H]
\centering
\vspace{-3mm}
\includegraphics[width=\linewidth]{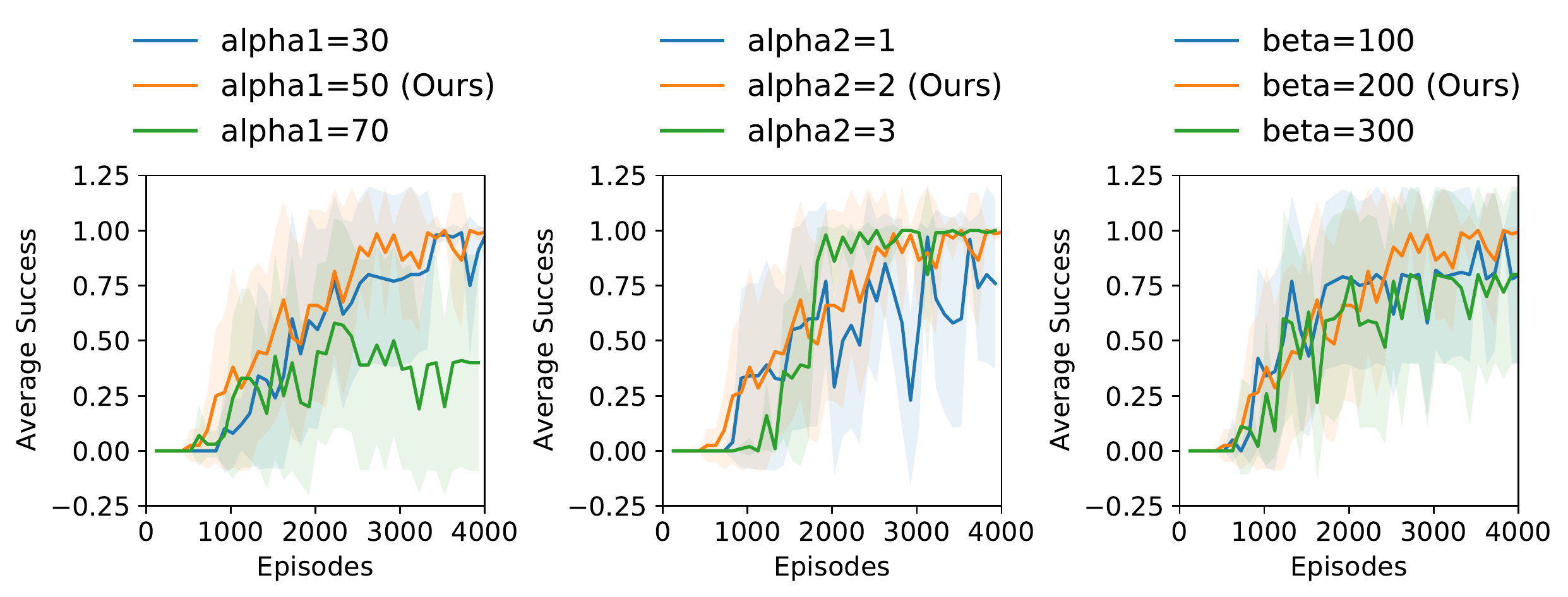}
\caption{\rebuttal{Training curves with different reward weights. For each plot, we train the policies by changing one of the weight terms to three different values.}}
\label{fig:weights}
\vspace{-3mm}
\end{figure}

\section{Implementation Details}
\label{appendix:details}
\vspace{-2mm}

\subsection{Simulation environment} 
\vspace{-2mm}

We build the simulation environment with Robosuite~\citep{zhu2020robosuite} which uses the MuJoCo simulator~\citep{todorov2012mujoco}. Each episode has a length of 40 timesteps which corresponds to 20 seconds of real time execution. At the beginning of each episode, we set the robot arm to an initial joint configuration with Gaussian noise in the joint angles of 0.02 rad. We use a box-shaped object in the simulation environment. The dimensions of the box are randomized in the ADR experiments. One important note on the simulator environment is the parameters of the MuJoCo solver. We notice that MuJoCo sometimes creates unrealistic contacts with the default solver. We reduce the simulation solver timestep from the default value of 0.002 to 0.001 and set the ``noslip iterations'' to 20 which significantly improved simulation quality on contacts.

\subsection{Grasp configurations}
\vspace{-2mm}

In this work, we focus on grasping large objects from the side because this is a task that may demonstrate the benefits of extrinsic dexterity. For single grasp experiments, a default grasp location is shown in Figure~\ref{fig:figure1}. In multi-grasp experiments, the grasps are sampled from a distribution shown in Figure~\ref{fig:multigrasp}. The grasps are sampled along the side of the box and they are 2 cm away from the edges. These grasp configurations are supposed to be the input to our proposed system, and could be replaced by other grasp generation methods. 

\subsection{Success rate calculation}
\vspace{-2mm}

In simulation, the success of the task is computed as $\mathbbm{1}(\Delta T<3 \, cm)\cdot \mathbbm{1}(\Delta \theta<10 \, deg) $ at the end of an episode. As defined in Section~\ref{task}, $\Delta T$ is the position difference between the end-effector and the target grasp and $\Delta \theta$ is the orientation difference. The success is defined in this way because we focus on reaching the desired grasp. One alternative is to evaluate the final grasping success by closing the gripper and lift the object. However, this will increase the simulation time during training. To confirm that the pose difference is a good proxy for the final grasping success, we evaluated a trained policy and verified that if the robot closes the gripper at the end of a successful episode according to the pose difference metric, it is able to lift the object $100\%$ of the time. 

For the real robot experiments, we evaluate success by closing the gripper and lifting the object; if the object was successfully lifted, we will mark it as a successful episode.

\subsection{Observation and action space}
\vspace{-2mm}

As mentioned in Section~\ref{method:rl}, the observation includes a target grasp configuration in the object frame ${}^O g$, the pose of the end-effector in the world frame ${}^W E$ and the object pose in the world frame ${}^W O$. One implementation detail is that we also include the pose of the end-effector in the object frame ${}^O E=({}^W O)^{-1} ({}^W E)$ because we found that it sometimes speeds up learning. Each pose is represented as a 3D translation vector and a 4D quaternion representation of the rotation. 

\begin{figure}[H]
    \centering
    \includegraphics[width=0.95\linewidth]{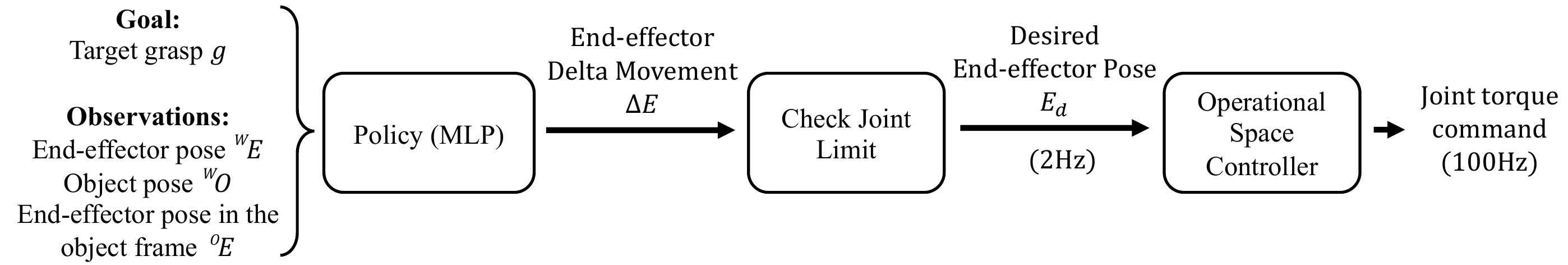}
    \caption{Outline of policy execution: Given the observation, the policy outputs an end-effector delta movement. If the desired pose is within the joint limit of the robot, it will be sent to the low-level controller which operates at a higher frequency.}
    \label{fig:policy}
    \vspace{-3mm}
\end{figure}

The action space of the policy is the delta pose of the end-effector $\Delta E$ in its local frame represented by a vector of translation $p\in \mathbb{R}^3$ and a 3D vector of rotation $q\in SO(3)$ with axis-angle representation. An outline of the policy execution pipeline is shown in Figure~\ref{fig:policy}. $\Delta E$ is then passed into a collision check function to form a desired pose $E_d$ which will be sent to a low-level controller. 

\subsection{Low-level controller}
\vspace{-2mm}

\textbf{Handling joint limit:} Although we may use nullspace in the operational space controller to avoid reaching joint limit, in practice, certain desired end-effector poses still reach joint limits that cannot be avoided by nullspace. Thus, we handling the joint limit in the following way. If the corresponding joint configuration of the desired pose is going to reach joint limits, we will overwrite the policy action to output the desired pose of the previous timestep to the low-level controller. In detail, we use the Jacobian $J$ to estimate the joint configuration of the desired pose:
\begin{equation}
\theta^{t+1}_{joints} = \theta^{t}_{joints} + J^{-1}\cdot \Delta E
\end{equation}
where $\theta_{joints}$ are the joint angles and $\Delta E$ is the output of the policy. If any joint in $\theta^{t+1}_{joints}$ is close to the limit, the low-level controller will use the previous desired pose $E_d$ instead.

\textbf{Parameters of the Operational Space Controller:} We use $K_p=300$ for position error, $K_p=30$ for orientation error, and $K_d=\sqrt{K_p}$. These values are chosen by making sure that the real robot is compliant enough to safely collide with the object and the bin without damage. In Figure~\ref{fig:ablations}, the baseline of ``High-gain OSC" uses $K_p=600$ for position error, $K_p=60$ for orientation error, and $K_d=\sqrt{K_p}$. This baseline with less compliance is not only slower to train in the simulation, but also not safe to execute on the real robot for our task which involves rich contacts and relies on environment constraints. During our initial experiments, with the high-gain OSC, the robot deforms the object and the bin surface.

\subsection{Multi-Grasp Training with Curriculum}
Here are more details on multi-grasp training. When the success rate of policy on a boundary case of the training range is above $0.8$, it will expand the range of grasps by $0.25$ (See Figure~\ref{fig:multigrasp} for parameterizations of the grasp configurations). For example, if the policy is currently training with grasps $[1, 2]$, and the success rate evaluated at grasp ID $1$ is above $0.8$, the new training range will be $[0.75, 2]$. This is following a similar procedure as Automatic Domain Randomization, but randomizing goals instead of simulation parameters.

\subsection{RL Training}
We use Soft Actor Critic~\citep{haarnoja2018soft} to train the RL policy with the impementation from rlkit (\url{https://github.com/rail-berkeley/rlkit}). \rebuttal{Hyperparameters for SAC training are included in Table~\ref{tab:hyperparameters}}. Since the task is conditioned on the target grasp as a goal, we use Hindsight Experience Replay~\citep{andrychowicz2017hindsight} for all the experiments with $60\%$ original goals and $40\%$ of the goals sampled from the same rollout. We compare the policies across 5 random seeds of each method and plot the average performance with standard deviation across seeds. We use 10 episodes for each evaluation setting.

\begin{table}[H]
\centering
\caption{\rebuttal{Hyperparameters for RL training.}}
\label{tab:hyperparameters}
\vspace{2mm}
\scalebox{0.9}{
\begin{tabular}{ll}
\hline
\textbf{Hyperparameters} & \textbf{Values} \\ \hline
Optimizer & Adam \\
Learning rate - Policy & 1e-3 \\
Leraning rate - Q-function & 5e-4 \\
Networks & {[}512, 512, 512{]} MLP \\
Batch size & 256 \\
Nonlinearity & ReLU \\
Soft target update ($\tau$) & 0.005 \\
Replay buffer size & 1e6 \\
Discount factor ($\gamma$) & 0.99 \\
HER rollout goals & 40\% \\ \hline
\end{tabular}}
\end{table}

\section{Automatic Domain Randomization}
\label{appendix:adr}

As discussed in Section~\ref{method:generalization}, we use Automatic Domain Randomization~\citep{openai2019solving} to improve policy generalization across environment variations. In ADR, the policy is first trained with an environment with very little randomization, and then we gradually expand the variations based on the evaluation performance. For a set of environment parameters $\lambda_i$, each $\lambda_i$ is sampled from a uniform distribution $\lambda_i \sim U(\phi_i^L, \phi_i^H)$ at the beginning of each episode. During training, the policy will be evaluated at these boundary values $\lambda_i=\phi_i^L$ or $\lambda_i=\phi_i^H$. If the performance is higher than a threshold, the boundary value will be expanded by an increment $\Delta$. For example, if the performance at $\lambda_i=\phi_i^H$ is higher than the threshold, the training distribution becomes $\lambda_i \sim U(\phi_i^L, \phi_i^H+\Delta)$ in the next iteration. Compared to directly training the policy with the entire variations, Automatic Domain Randomization can reduce the need of manually tuning a suitable range of variations for each environment parameter.

Table~\ref{tab:parameters} summarized the simulation parameters in the experiment. All the parameters are uniformly sampled from these ranges at the beginning of each episode. The ranges of the parameters start from a single initial value and gradually expand to a wider range according to the pre-specific increment step $+\Delta$ on the upper bound and the decrement step $-\Delta$ at the lower bound. 

\begin{table}[htb]
\vspace{-2mm}
\caption{Simulation parameters in Automatic Domain Randomization}
\vspace{2mm}
\label{tab:parameters}
\centering
  \scalebox{0.9}{
\begin{tabular}{l|cccc}
\hline
                             & Initial Value & $+\Delta$ & $-\Delta$ & Final Range        \\ \hline
Object size x (m)                & 0.15          & 0.01           & -0.01          & {[}0.14, 0.16{]}   \\
Object size z (m)               & 0.05          & 0.01           & -0.01          & {[}0.04, 0.06{]}   \\
Table friction               & 0.3           & 0.1            & -0.1           & {[}0.1, 0.5{]}     \\
Gripper friction             & 3             & /              & -1             & {[}2, 3{]}         \\
Object Density ($g/m^3$)               & 86            & 86             & 43             & {[}43, 172{]}      \\
Action translation scale (m) & 0.03          & /              & -0.005         & {[}0.02, 0.03{]}   \\
Action rotation scale (rad)    & 0.2           & /              & -0.05          & {[}0.1, 0.2{]}     \\
Initial distance to wall (m)     & 0             & 0.01           & /              & {[}0, 0.02{]}      \\
Table offset x (m)              & 0.5           & 0.01           & -0.01          & {[}0.48, 0.52{]}   \\
Table offset z (m)              & 0.07          & 0.01           & 0.01           & {[}0.055, 0.075{]} \\ \hline
\end{tabular}}
\end{table}

\begin{figure}[H]
    \centering
    \begin{subfigure}{\textwidth}
    \centering
    \includegraphics[width=0.6\linewidth]{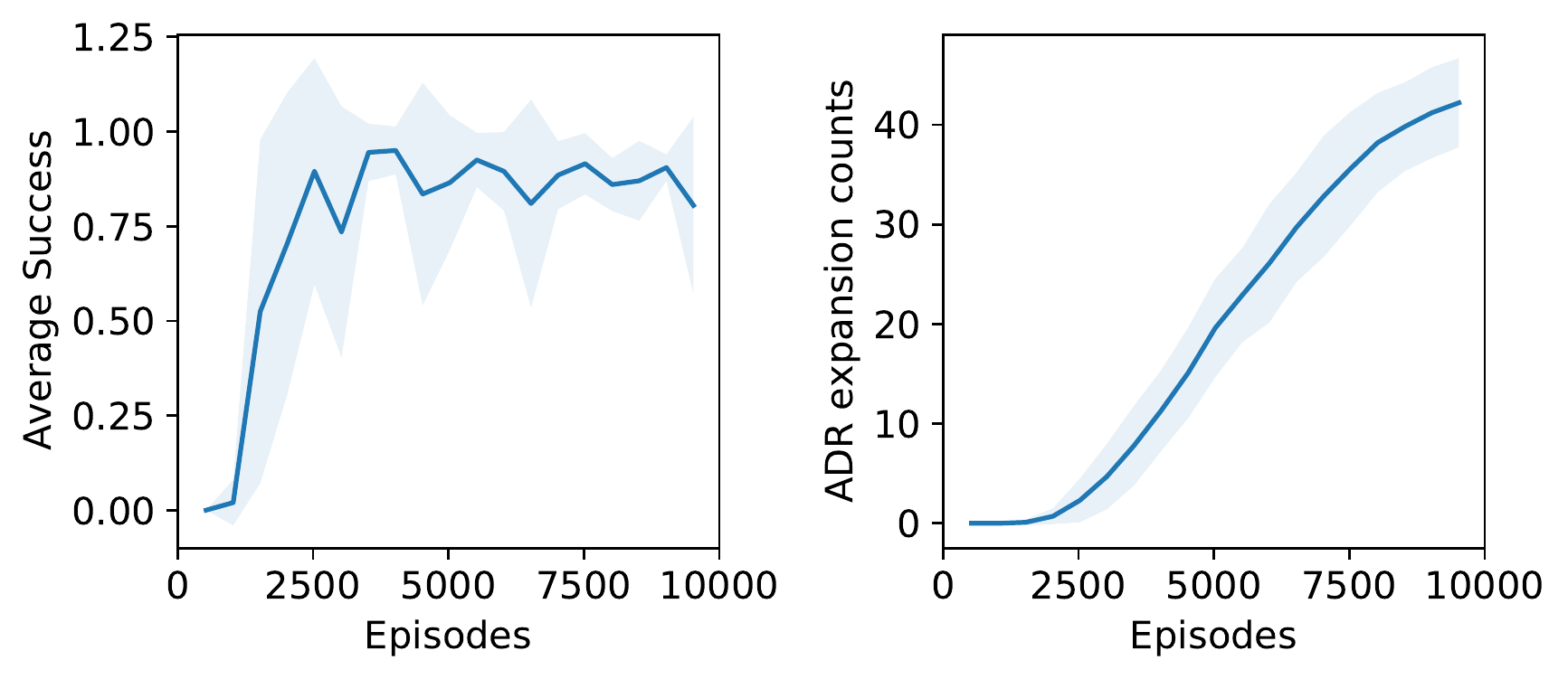}  
    \caption{\rebuttal{Overall training performance of the ADR policies: Success rate over the entire training range (left) and total number of expanded parameter boundaries.}}
    \label{fig:multigraspfront}
    \end{subfigure}
    \hfill
    \vspace{3mm}
    \begin{subfigure}{\textwidth}
    \centering
    \includegraphics[width=0.9\linewidth]{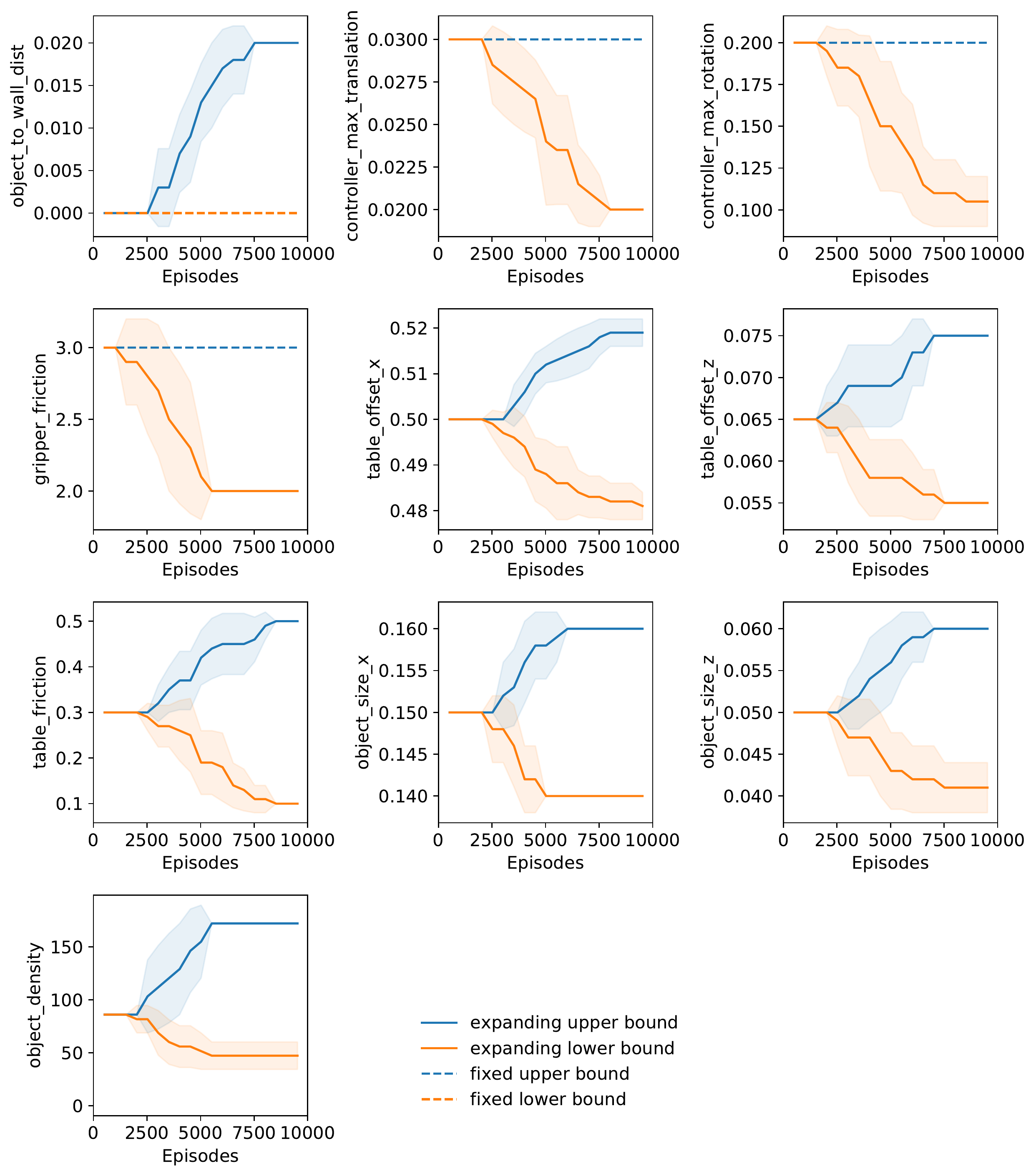}
    \caption{\rebuttal{Training progress of individual ADR parameters. Each plot for the physical parameters has two curves indicating the upper and lower bound of the expanded training range. Dashed lines indicate fixed parameter boundaries where we do not intent to expand.}}
    \label{fig:multigraspside}
    \end{subfigure}

\caption{\rebuttal{Training curves for the ADR policies.}}
\label{fig:adr}
\vspace{-2mm}
\end{figure}

\rebuttal{We include the training plots of the ADR policies in Figure~\ref{fig:adr}. Dashed lines in Figure~\ref{fig:adr} indicate fixed parameter boundaries where we do not intent to expand. The final ranges are used when we sample 100 environments for evaluation in Section~\ref{exp:generalization}.} 

\section{Real robot experiment}
\label{appendix:realrobot}
In this section, we include more details and discussion for the real robot experiments. Quantitative results can be found on the website~\footnote{\url{https://sites.google.com/view/grasp-ungraspable}} where we include all the videos for the real robot experiments, video examples of failure cases, recovery behaviors and ICP results.

\subsection{Implementation details}
\begin{figure}[H]
    \centering
    \includegraphics[width=0.6\linewidth]{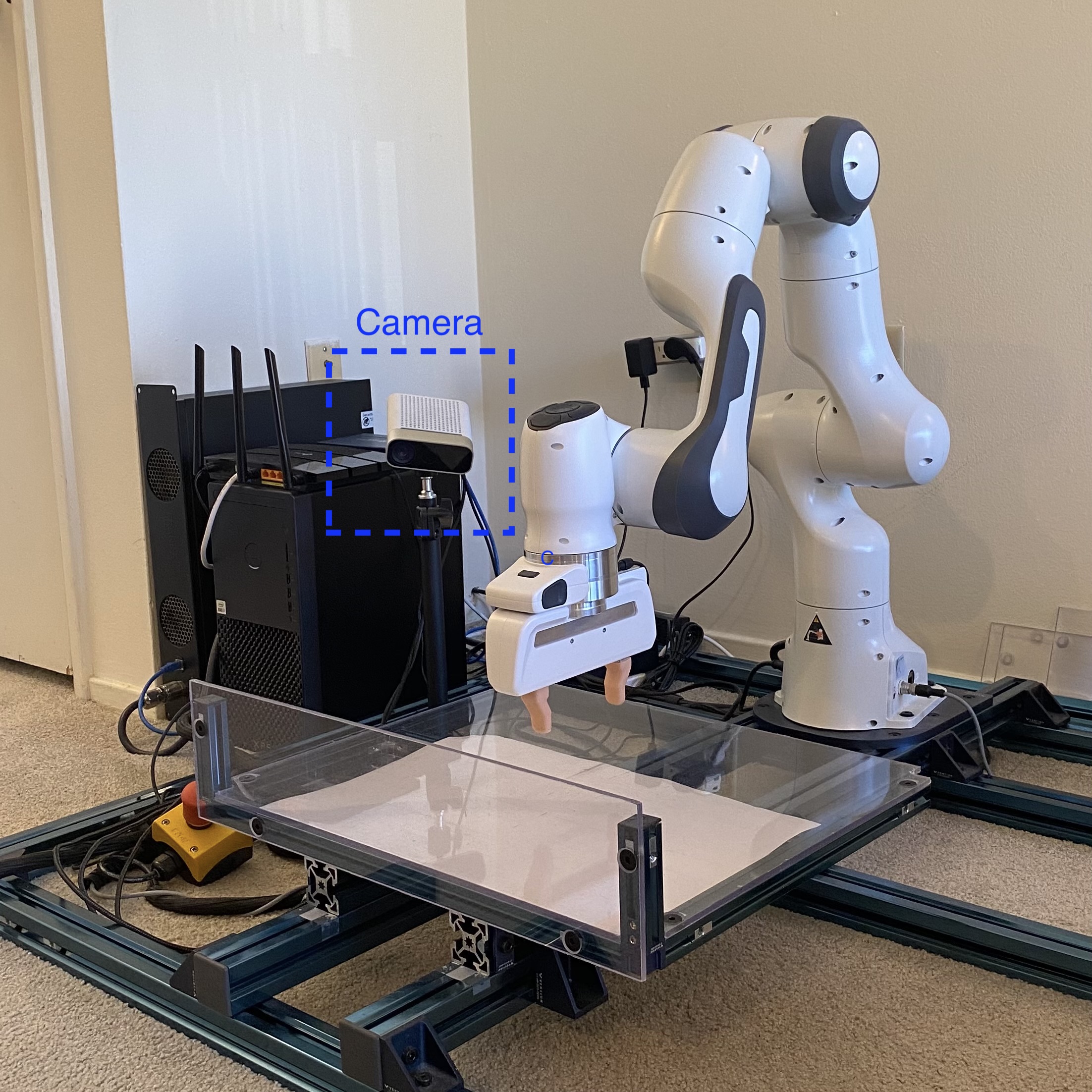}
    \caption{\rebuttal{Robot setup. We use one Azure Kinect camera for object pose estimation.}}
    \label{fig:setup}
    \vspace{-2mm}
\end{figure}

\rebuttal{The robot setup is shown in Figure~\ref{fig:setup}}. The code for controlling the real robot is built on top of FrankaPy~\citep{zhang2020modular}. The policies are trained in the simulator and zero-shot transferred to a physical Franka Emika Panda robot. For the real robot experiments, we train a policy in the XZ plane from the side view to reduce the sim2real gap of the policy, since the motion is mostly in the XZ plane for the side grasp.

\textbf{Sim2Real gap of the low-level controller:} We observe a noticeable sim2real gap on the low-level controller when deploying the policy. The same command of moving the end-effector to a certain pose in free space may not have the same resulting movement. This is a combination of two factors: First, there is a significant discrepancy between the robot model in simulation and the real robot. The real robot has more damping and friction on the joints. Second, we use a compliant controller for this task, which is more susceptible to the noise in the system. As a result, the real robot always executes a smaller delta movement than the simulator. To compensate for this sim2real gap, we slightly increase the action scale and reduce the policy execution rate from 2Hz to 1Hz. Both of these changes will allow the real robot to compensate for the smaller movement caused by the damping and friction of its joints. Note that the sim2real gap still exists after these changes. However, the remaining gap could be further compensated by using a closed-loop policy. During our experiments, we first use the default object Box-0 to tune the controller until we observe several successes in a row. After that, we keep the same controller setting for the entire evaluation process.

\begin{figure}[H]
    \centering
    \includegraphics[width=0.9\linewidth]{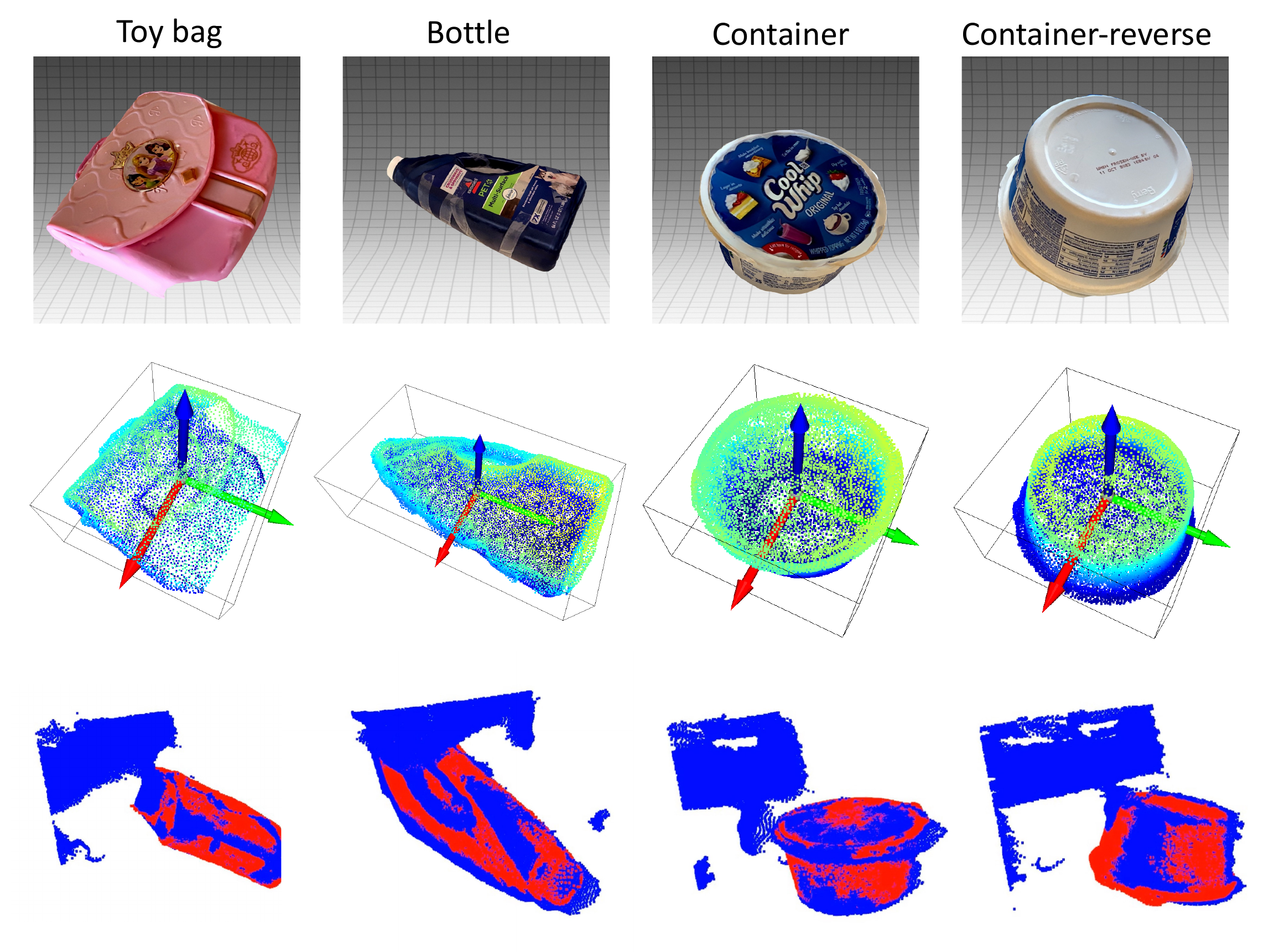}
    \caption{Illustrations of pose estimation pipeline for the non-box objects. The top row shows the scanned object model. The middle row shows bounding box calculation and pose definition. The last row shows an example of ICP.}
    \label{fig:object_pose}
    \vspace{-2mm}
\end{figure}

\textbf{Defining object pose:} The pose of a box can simply be defined at the center of its volume and with the axes defined parallel to the edges. For non-box object, we define the pose to be the center of its bounding box. We scan the non-box objects into point clouds with the Qlone app on the phone (Figure~\ref{fig:object_pose} top row). To obtain the bounding box, we first run Principle Component Analysis of the scanned object to get the principle axes. Then, we take the min and max values along the axes to get the dimension of the bounding box. The axes are then aligned to global axes based on the initial pose (Figure~\ref{fig:object_pose} middle row).

\begin{figure}[H]
    \centering
    \includegraphics[width=0.7\linewidth]{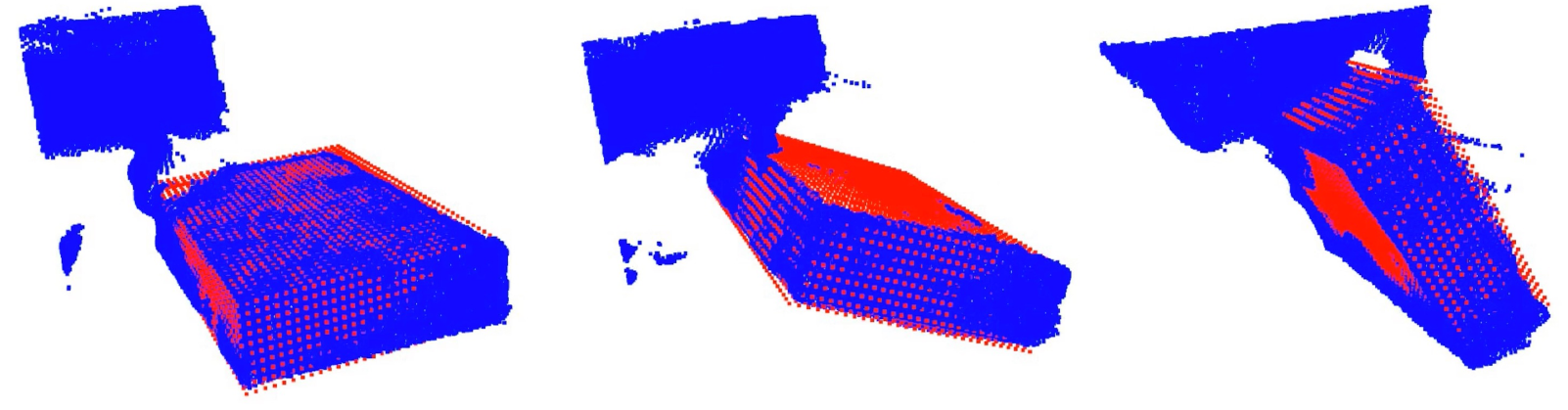}
    \caption{Examples of pose estimation with ICP during an episode. The blue points are the observed point cloud from the camera. The red points are the object template that matches to the observed point cloud using ICP.}
    \label{fig:icp}
\end{figure}

\textbf{Pose estimation with Iterative Closest Point:} To get the pose of the object as the observation of the policy, we use Iterative Closest Point (ICP) which matches the current point cloud to a template point cloud of the object~\citep{rusinkiewicz2001efficient}. \rebuttal{We use the implementation from Open3D}. For box objects, we simply create a box shape template with measured size. For non-box objects, we use the scanned object point clouds as mentioned above. Figure~\ref{fig:icp} shows an example of the results from ICP across an episode. Figure~\ref{fig:object_pose} includes examples of ICP results for non-box objects potentially with partial point cloud. More visualizations of ICP results can be found on the website.

\subsection{More information on the objects}
\vspace{-2mm}
To emphasize the diversity of the objects and demonstrate the generalization capability of the policy, we include more descriptions on the objects in this section. In Table~\ref{tab:objects}, we highlight the object properties that are out of the ADR training distribution in bold. Box-0 is the default object that we used to calibrate the simulator and to tune the low-level controller. 

\begin{table}[H]
  \caption{Real robot evaluations with more object information. We highlight the out-of-distribution aspect of the object properties in bold.}
  \vspace{3mm}
  \label{tab:objects}
  \centering
  \scalebox{0.85}{\begin{tabular}{lcccccc}
    \hline\multicolumn{1}{l}{\bf Object-ID}
    &\multicolumn{1}{l}{\bf Box?}
    &\multicolumn{1}{c}{\bf Surface Material}
    &\multicolumn{1}{c}{\bf \makecell{Bounding Box \\Dimension (cm)}}
    &\multicolumn{1}{c}{\bf \makecell{Weight \\(g)}}
    &\multicolumn{1}{c}{\bf \makecell{Success \\ w/o ADR}}
    &\multicolumn{1}{c}{\bf \makecell{Success \\ w/ ADR}}\\
    
    \hline
    Box-0	            &Yes &Cardboard &(15.0, 20.0, 5.0)   &   128   &   9/10    &   9/10\\
    Box-0 + 4 erasers	&Yes &Cardboard &(15.0, 20.0, 5.0)   &   237   &   6/10   &   10/10\\
    Box-0 + 8 erasers	&Yes &Cardboard &(15.0, 20.0, 5.0)   &   \textbf{345}   &   3/10    &   4/10\\
    Box-1	            &Yes &Cardboard &(15.4, 29.2, 5.8)   &   130   &   5/10    &   8/10\\
    Box-2	            &Yes &Cardboard & \textbf{(15.3, 22.2, 7.4)}   &   113   &   2/10    &   9/10\\
    Box-3	            &Yes & \textbf{Cardboard with tape} & \textbf{(16.5, 24.5, 5.2)}   &   \textbf{50}    &   0/10    &   7/10\\
    Toy Bag	            &Almost &\textbf{Silicone} & \textbf{(16.6, 14.5, 7.1)}   &   203    &   8/10    &   7/10\\
    Bottle	            &\textbf{No} &\textbf{Plastic} & \textbf{(16.3, 28.8, 9.0)}   &   112    &   0/10    &   8/10\\
    Container	        &\textbf{No} &\textbf{Plastic} & \textbf{(14.7, 14.7, 8.1)}   &   \textbf{61}    &   0/10   &   10/10\\
    Container-reverse   &\textbf{No} &\textbf{Plastic} & \textbf{(14.7, 14.7, 8.1)}   &   \textbf{61}    &   0/10    &   6/10\\\hline
    Average	           & &   &     &   & $33\%$     &   $78\%$ \\\hline
  \end{tabular}}
\end{table}

The policy trained with ADR can generalize across physical properties such as weight and surface friction. We stress test Box-0 with additional weights by putting four or eight erasers inside of the box. The erasers can move in the box during execution, which is not modeled in simulation. Although we do not have access to the true friction coefficient between the object and the table, the difference in surface friction results in qualitatively different behavior of the object even among the cardboard boxes. For example, Box-3 has tape on its surface which has much higher friction than the others cardboard boxes. It tends to stick to the wall during execution. The toy bag has a similar cross section as the box but the material is very different.

We also evaluate the policies with objects that are not similar to a box shape including a bottle and a container. Due to the difference in shape, both objects result in different dynamics during execution. In addition, with the same container object, starting it from different initial poses will also lead to different object pose distribution. Videos can be found on the website. Nonetheless, the policy trained with ADR shows reasonable generalization across these non-box objects.

\subsection{Failure cases}

In this section, we include discussions on the failure cases of the evaluation. We categorize the failure cases into the following categories and discuss the potential reasons:

A failure case that happens before the initial contact:
\begin{itemize}
    \item \textbf{Missing initial contact}: The robot is not able to reach the initial contact of the object to rotate it. This is mostly due to the noise in pose estimation and the variations in object dimension.
\end{itemize}
    
Failure cases that happen during the rotation:
\begin{itemize}
    \item \textbf{Object drops during rotation}: The object drops to the table during rotation. One potential reason for this failure case is that the finger slips on the object during rotation. Another potential reason for this failure case is the insufficient rotation of the low-level controller due to the sim2real gap (See Section~\ref{appendix:realrobot} - Sim2Real gap of the low-level controller). In the ``dropping'' strategy, the policy is supposed to rotate object and then let it drop on the bottom finger. Before the dropping happens, the gripper needs to be rotated until the bottom finger is below the object. Otherwise, the bottom finger will not be able to catch the object and the object directly drops to the table.
    \item \textbf{Repeated rotation}: The robot repeatedly rotates and drops the object. This is different from the previous failure case because the robot moves down with the object at the same time. Our hypothesis for this failure case is that the policy gets stuck in a loop in the MDP.
    \item \textbf{Joint limit}: The robot hits a joint limit and the policy gets stuck at the joint limit.
\end{itemize}

Failure cases that happen after the rotation:
\begin{itemize}
    \item \textbf{Unexpected object dynamics}: When the robot rotates the object, the object might move in unexpected ways. This mostly happens for the non-box objects.
    \item \textbf{Stop reaching}: Following the ``standing'' strategy, the policy successfully rotates the object to a stable pose on the side of the object. However, it cannot reach the final grasping pose. The gripper tries to move down to reach the pose but it collides with the object due to the unexpected object dimension.
    \item \textbf{Timeout}: Since we use a fixed episode length during evaluation, sometimes the policy does not have enough time to finish the task although it is very close to a success. This happens when the policy spends time to recover from some failed attempts at the beginning of the episode.
\end{itemize}

Videos on these failure cases can be found on the website. We summarize the counts of the failure cases in Table~\ref{tab:failure-1} and Table~\ref{tab:failure-2}. The most common failure case for the policy trained with ADR is the repeated rotation. For the policy trained without ADR, the most common failure case is missing the initial contact. Comparing the percentage of the failure cases between Table~\ref{tab:failure-1} and Table~\ref{tab:failure-2}, we observe that percentage of missing the initial contact and the percentage of stopping the reaching motion drops drastically when the policy is trained with ADR because the policy is more robust to variations in shape and dimensions.
\input{table-failure}

\end{appendices}

%% file: table-failure.tex

\begin{table}[H]

\caption{Failure cases for \textbf{Policy w/ ADR} during real robot evaluation. The most common failures include dropping the object during rotation, repeated rotation, and unexpected object dynamics.}
\label{tab:failure-1}
\vspace{3mm}
\centering
\scalebox{0.9}{
\begin{tabular}{lccccccc}
\hline
\textbf{} & \textbf{\begin{tabular}[c]{@{}c@{}}Initial\\ contact\end{tabular}} & \textbf{\begin{tabular}[c]{@{}c@{}}Object\\ drops\end{tabular}} & \textbf{\begin{tabular}[c]{@{}c@{}}Repeated\\ rotation\end{tabular}} & \textbf{\begin{tabular}[c]{@{}c@{}}Joint\\ limit\end{tabular}} & \textbf{\begin{tabular}[c]{@{}c@{}}Unexpected\\ dynamics\end{tabular}} & \textbf{\begin{tabular}[c]{@{}c@{}}Stop\\ reaching\end{tabular}} & \textbf{Timeout} \\ \hline
\textbf{Box-0} & 0 & 0 & 1 & 0 & 0 & 0 & 0 \\
\textbf{Box-0 + 4 erasers} & 0 & 0 & 0 & 0 & 0 & 0 & 0 \\
\textbf{Box-0 + 8 erasers} & 0 & 3 & 3 & 0 & 0 & 0 & 0 \\
\textbf{Box-1} & 0 & 1 & 1 & 0 & 0 & 0 & 0 \\
\textbf{Box-2} & 0 & 0 & 0 & 1 & 0 & 0 & 0 \\
\textbf{Box-3} & 0 & 0 & 2 & 0 & 0 & 0 & 0 \\
\textbf{Toy Bag} & 0 & 2 & 1 & 0 & 0 & 0 & 0 \\
\textbf{Bottle} & 1 & 0 & 0 & 0 & 1 & 0 & 0 \\
\textbf{Container} & 0 & 0 & 0 & 0 & 0 & 0 & 0 \\
\textbf{Container-reverse} & 0 & 0 & 0 & 0 & 4 & 0 & 0 \\ \hline
\textbf{Total} & \textbf{1} & \textbf{6} & \textbf{8} & \textbf{1} & \textbf{5} & \textbf{0} & \textbf{0} \\
\textbf{Percentage} & \textbf{4.8\%} & \textbf{28.5\%} & \textbf{38.1\%} & \textbf{4.8\%} & \textbf{23.8\%} & \textbf{0.0\%} & \textbf{0.0\%} \\ \hline
\end{tabular}}
\end{table}

\begin{table}[H]
\caption{Failure cases for \textbf{Policy w/o ADR} during real robot evaluation. The most common failures include missing the initial contact, repeated rotation and unexpected object dynamics.}
\label{tab:failure-2}
\vspace{3mm}
\centering
\scalebox{0.9}{
\begin{tabular}{lccccccc}
\hline
\textbf{} & \textbf{\begin{tabular}[c]{@{}c@{}}Initial\\ contact\end{tabular}} & \textbf{\begin{tabular}[c]{@{}c@{}}Object\\ drops\end{tabular}} & \textbf{\begin{tabular}[c]{@{}c@{}}Repeated\\ rotation\end{tabular}} & \textbf{\begin{tabular}[c]{@{}c@{}}Joint\\ limit\end{tabular}} & \textbf{\begin{tabular}[c]{@{}c@{}}Unexpected\\ dynamics\end{tabular}} & \textbf{\begin{tabular}[c]{@{}c@{}}Stop\\ reaching\end{tabular}} & \textbf{Timeout} \\ \hline
\textbf{Box-0} & 0 & 0 & 0 & 0 & 0 & 0 & 1 \\
\textbf{Box-0 + 4 erasers} & 1 & 0 & 2 & 0 & 0 & 0 & 1 \\
\textbf{Box-0 + 8 erasers} & 0 & 1 & 3 & 0 & 0 & 2 & 1 \\
\textbf{Box-1} & 2 & 0 & 2 & 0 & 1 & 0 & 0 \\
\textbf{Box-2} & 1 & 0 & 1 & 0 & 0 & 6 & 0 \\
\textbf{Box-3} & 10 & 0 & 0 & 0 & 0 & 0 & 0 \\
\textbf{Toy Bag} & 0 & 2 & 0 & 0 & 0 & 0 & 1 \\
\textbf{Bottle} & 6 & 0 & 0 & 0 & 4 & 0 & 0 \\
\textbf{Container} & 0 & 0 & 2 & 0 & 8 & 0 & 0 \\
\textbf{Container-reverse} & 1 & 1 & 6 & 0 & 1 & 0 & 1 \\ \hline
\textbf{Total} & \textbf{21} & \textbf{4} & \textbf{16} & \textbf{0} & \textbf{14} & \textbf{8} & \textbf{5} \\ 
\textbf{Percentage} & \textbf{30.9\%} & \textbf{5.9\%} & \textbf{23.5\%} & \textbf{0.0\%} & \textbf{20.6\%} & \textbf{11.8\%} & \textbf{7.3\%} \\ \hline
\end{tabular}}
\end{table}

%% file: main.bbl
\begin{thebibliography}{41}
\providecommand{\natexlab}[1]{#1}
\providecommand{\url}[1]{\texttt{#1}}
\expandafter\ifx\csname urlstyle\endcsname\relax
  \providecommand{\doi}[1]{doi: #1}\else
  \providecommand{\doi}{doi: \begingroup \urlstyle{rm}\Url}\fi

\bibitem[Dafle et~al.(2014)Dafle, Rodriguez, Paolini, Tang, Srinivasa, Erdmann,
  Mason, Lundberg, Staab, and Fuhlbrigge]{dafle2014extrinsic}
N.~C. Dafle, A.~Rodriguez, R.~Paolini, B.~Tang, S.~S. Srinivasa, M.~Erdmann,
  M.~T. Mason, I.~Lundberg, H.~Staab, and T.~Fuhlbrigge.
\newblock Extrinsic dexterity: In-hand manipulation with external forces.
\newblock In \emph{2014 IEEE International Conference on Robotics and
  Automation (ICRA)}, pages 1578--1585. IEEE, 2014.

\bibitem[Chavan-Dafle and Rodriguez(2017)]{chavandafle2017samplingbased}
N.~Chavan-Dafle and A.~Rodriguez.
\newblock Sampling-based planning of in-hand manipulation with external pushes,
  2017.

\bibitem[Hou et~al.(2020)Hou, Jia, and Mason]{hou2020manipulation}
Y.~Hou, Z.~Jia, and M.~Mason.
\newblock Manipulation with shared grasping.
\newblock In \emph{Robotics: Science and Systems}, 2020.

\bibitem[Hou et~al.(2018)Hou, Jia, and Mason]{8462834}
Y.~Hou, Z.~Jia, and M.~T. Mason.
\newblock Fast planning for 3d any-pose-reorienting using pivoting.
\newblock In \emph{2018 IEEE International Conference on Robotics and
  Automation (ICRA)}, pages 1631--1638, 2018.
\newblock \doi{10.1109/ICRA.2018.8462834}.

\bibitem[Chavan-Dafle et~al.(2019)Chavan-Dafle, Holladay, and
  Rodriguez]{chavandafle2019inhand}
N.~Chavan-Dafle, R.~Holladay, and A.~Rodriguez.
\newblock In-hand manipulation via motion cones, 2019.

\bibitem[Cheng et~al.(2021{\natexlab{a}})Cheng, Huang, Hou, and
  Mason]{cheng2021contact2d}
X.~Cheng, E.~Huang, Y.~Hou, and M.~T. Mason.
\newblock Contact mode guided sampling-based planning for quasistatic dexterous
  manipulation in 2d.
\newblock In \emph{2021 IEEE International Conference on Robotics and
  Automation (ICRA)}, pages 6520--6526. IEEE, 2021{\natexlab{a}}.

\bibitem[Cheng et~al.(2021{\natexlab{b}})Cheng, Huang, Hou, and
  Mason]{cheng2021contact3d}
X.~Cheng, E.~Huang, Y.~Hou, and M.~T. Mason.
\newblock Contact mode guided motion planning for quasidynamic dexterous
  manipulation in 3d.
\newblock \emph{arXiv preprint arXiv:2105.14431}, 2021{\natexlab{b}}.

\bibitem[Mousavian et~al.(2019)Mousavian, Eppner, and
  Fox]{mousavian2019graspnet}
A.~Mousavian, C.~Eppner, and D.~Fox.
\newblock 6-dof graspnet: Variational grasp generation for object manipulation.
\newblock In \emph{International Conference on Computer Vision (ICCV)}, 2019.

\bibitem[Murali et~al.(2020)Murali, Mousavian, Eppner, Paxton, and
  Fox]{murali20206dof}
A.~Murali, A.~Mousavian, C.~Eppner, C.~Paxton, and D.~Fox.
\newblock 6-dof grasping for target-driven object manipulation in clutter,
  2020.

\bibitem[Wang et~al.(2020)Wang, Xiang, and Fox]{wang2020manipulation}
L.~Wang, Y.~Xiang, and D.~Fox.
\newblock Manipulation trajectory optimization with online grasp synthesis and
  selection.
\newblock In \emph{Robotics: Science and Systems (RSS)}, 2020.

\bibitem[Sun et~al.(2020)Sun, Yuan, Hu, Yang, and Li]{sun2020learning}
Z.~Sun, K.~Yuan, W.~Hu, C.~Yang, and Z.~Li.
\newblock Learning pregrasp manipulation of objects from ungraspable poses,
  2020.

\bibitem[Chang et~al.(2010)Chang, Srinivasa, and Pollard]{chang2010planning}
L.~Y. Chang, S.~S. Srinivasa, and N.~S. Pollard.
\newblock Planning pre-grasp manipulation for transport tasks.
\newblock In \emph{2010 IEEE International Conference on Robotics and
  Automation}, pages 2697--2704. IEEE, 2010.

\bibitem[Hang et~al.(2019)Hang, Morgan, and Dollar]{8610166}
K.~Hang, A.~S. Morgan, and A.~M. Dollar.
\newblock Pre-grasp sliding manipulation of thin objects using soft, compliant,
  or underactuated hands.
\newblock \emph{IEEE Robotics and Automation Letters}, 4\penalty0 (2):\penalty0
  662--669, 2019.
\newblock \doi{10.1109/LRA.2019.2892591}.

\bibitem[OpenAI et~al.(2019)OpenAI, Akkaya, Andrychowicz, Chociej, Litwin,
  McGrew, Petron, Paino, Plappert, Powell, Ribas, Schneider, Tezak, Tworek,
  Welinder, Weng, Yuan, Zaremba, and Zhang]{openai2019solving}
OpenAI, I.~Akkaya, M.~Andrychowicz, M.~Chociej, M.~Litwin, B.~McGrew,
  A.~Petron, A.~Paino, M.~Plappert, G.~Powell, R.~Ribas, J.~Schneider,
  N.~Tezak, J.~Tworek, P.~Welinder, L.~Weng, Q.~Yuan, W.~Zaremba, and L.~Zhang.
\newblock Solving rubik's cube with a robot hand, 2019.

\bibitem[Eppner and Brock(2017)]{8202168}
C.~Eppner and O.~Brock.
\newblock Visual detection of opportunities to exploit contact in grasping
  using contextual multi-armed bandits.
\newblock In \emph{2017 IEEE/RSJ International Conference on Intelligent Robots
  and Systems (IROS)}, pages 273--278, 2017.
\newblock \doi{10.1109/IROS.2017.8202168}.

\bibitem[Shimoga(1996)]{shimoga1996robot}
K.~B. Shimoga.
\newblock Robot grasp synthesis algorithms: A survey.
\newblock \emph{The International Journal of Robotics Research}, 15\penalty0
  (3):\penalty0 230--266, 1996.

\bibitem[Nguyen(1988)]{nguyen1988constructing}
V.-D. Nguyen.
\newblock Constructing force-closure grasps.
\newblock \emph{The International Journal of Robotics Research}, 7\penalty0
  (3):\penalty0 3--16, 1988.

\bibitem[Pinto and Gupta(2016)]{pinto2016supersizing}
L.~Pinto and A.~Gupta.
\newblock Supersizing self-supervision: Learning to grasp from 50k tries and
  700 robot hours.
\newblock In \emph{2016 IEEE international conference on robotics and
  automation (ICRA)}, pages 3406--3413. IEEE, 2016.

\bibitem[Bohg et~al.(2013)Bohg, Morales, Asfour, and Kragic]{bohg2013data}
J.~Bohg, A.~Morales, T.~Asfour, and D.~Kragic.
\newblock Data-driven grasp synthesis—a survey.
\newblock \emph{IEEE Transactions on Robotics}, 30\penalty0 (2):\penalty0
  289--309, 2013.

\bibitem[Murali et~al.(2020)Murali, Liu, Marino, Chernova, and
  Gupta]{murali2020object}
A.~Murali, W.~Liu, K.~Marino, S.~Chernova, and A.~Gupta.
\newblock Same object, different grasps: Data and semantic knowledge for
  task-oriented grasping, 2020.

\bibitem[Vahrenkamp et~al.(2010)Vahrenkamp, Do, Asfour, and Dillmann]{5509377}
N.~Vahrenkamp, M.~Do, T.~Asfour, and R.~Dillmann.
\newblock Integrated grasp and motion planning.
\newblock In \emph{2010 IEEE International Conference on Robotics and
  Automation}, pages 2883--2888, 2010.
\newblock \doi{10.1109/ROBOT.2010.5509377}.

\bibitem[Fontanals et~al.(2014)Fontanals, Dang-Vu, Porges, Rosell, and
  Roa]{7041469}
J.~Fontanals, B.-A. Dang-Vu, O.~Porges, J.~Rosell, and M.~A. Roa.
\newblock Integrated grasp and motion planning using independent contact
  regions.
\newblock In \emph{2014 IEEE-RAS International Conference on Humanoid Robots},
  pages 887--893, 2014.
\newblock \doi{10.1109/HUMANOIDS.2014.7041469}.

\bibitem[Wang et~al.(2021)Wang, Xiang, Yang, Mousavian, and
  Fox]{wang2021goalauxiliary}
L.~Wang, Y.~Xiang, W.~Yang, A.~Mousavian, and D.~Fox.
\newblock Goal-auxiliary actor-critic for 6d robotic grasping with point
  clouds, 2021.

\bibitem[Kalashnikov et~al.(2018)Kalashnikov, Irpan, Pastor, Ibarz, Herzog,
  Jang, Quillen, Holly, Kalakrishnan, Vanhoucke, and
  Levine]{kalashnikov2018qtopt}
D.~Kalashnikov, A.~Irpan, P.~Pastor, J.~Ibarz, A.~Herzog, E.~Jang, D.~Quillen,
  E.~Holly, M.~Kalakrishnan, V.~Vanhoucke, and S.~Levine.
\newblock Qt-opt: Scalable deep reinforcement learning for vision-based robotic
  manipulation, 2018.

\bibitem[Song et~al.(2020)Song, Zeng, Lee, and Funkhouser]{song2020grasping}
S.~Song, A.~Zeng, J.~Lee, and T.~Funkhouser.
\newblock Grasping in the wild: Learning 6dof closed-loop grasping from
  low-cost demonstrations.
\newblock \emph{Robotics and Automation Letters}, 2020.

\bibitem[King et~al.(2013)King, Klingensmith, Dellin, Dogar, Velagapudi,
  Pollard, and Srinivasa]{King-RSS-13}
J.~King, M.~Klingensmith, C.~Dellin, M.~Dogar, P.~Velagapudi, N.~Pollard, and
  S.~Srinivasa.
\newblock Pregrasp manipulation as trajectory optimization.
\newblock In \emph{Proceedings of Robotics: Science and Systems}, Berlin,
  Germany, June 2013.
\newblock \doi{10.15607/RSS.2013.IX.015}.

\bibitem[Chen et~al.(2021)Chen, Xu, and Agrawal]{chen2021system}
T.~Chen, J.~Xu, and P.~Agrawal.
\newblock A system for general in-hand object re-orientation.
\newblock \emph{Conference on Robot Learning}, 2021.

\bibitem[Nagabandi et~al.(2020)Nagabandi, Konolige, Levine, and
  Kumar]{nagabandi2020deep}
A.~Nagabandi, K.~Konolige, S.~Levine, and V.~Kumar.
\newblock Deep dynamics models for learning dexterous manipulation.
\newblock In \emph{Conference on Robot Learning}, pages 1101--1112. PMLR, 2020.

\bibitem[Lee et~al.(2021)Lee, Devin, Zhou, Lampe, Bousmalis, Springenberg,
  Byravan, Abdolmaleki, Gileadi, Khosid, et~al.]{lee2021beyond}
A.~X. Lee, C.~M. Devin, Y.~Zhou, T.~Lampe, K.~Bousmalis, J.~T. Springenberg,
  A.~Byravan, A.~Abdolmaleki, N.~Gileadi, D.~Khosid, et~al.
\newblock Beyond pick-and-place: Tackling robotic stacking of diverse shapes.
\newblock In \emph{5th Annual Conference on Robot Learning}, 2021.

\bibitem[Tobin et~al.(2017)Tobin, Fong, Ray, Schneider, Zaremba, and
  Abbeel]{tobin2017domain}
J.~Tobin, R.~Fong, A.~Ray, J.~Schneider, W.~Zaremba, and P.~Abbeel.
\newblock Domain randomization for transferring deep neural networks from
  simulation to the real world, 2017.

\bibitem[Schaul et~al.(2015)Schaul, Horgan, Gregor, and
  Silver]{schaul2015universal}
T.~Schaul, D.~Horgan, K.~Gregor, and D.~Silver.
\newblock Universal value function approximators.
\newblock In \emph{International conference on machine learning}, pages
  1312--1320. PMLR, 2015.

\bibitem[Rusinkiewicz and Levoy(2001)]{rusinkiewicz2001efficient}
S.~Rusinkiewicz and M.~Levoy.
\newblock Efficient variants of the icp algorithm.
\newblock In \emph{Proceedings third international conference on 3-D digital
  imaging and modeling}, pages 145--152. IEEE, 2001.

\bibitem[Khatib(1987)]{khatib1987unified}
O.~Khatib.
\newblock A unified approach for motion and force control of robot
  manipulators: The operational space formulation.
\newblock \emph{IEEE Journal on Robotics and Automation}, 3\penalty0
  (1):\penalty0 43--53, 1987.

\bibitem[Mart{\'\i}n-Mart{\'\i}n et~al.(2019)Mart{\'\i}n-Mart{\'\i}n, Lee,
  Gardner, Savarese, Bohg, and Garg]{martin2019variable}
R.~Mart{\'\i}n-Mart{\'\i}n, M.~A. Lee, R.~Gardner, S.~Savarese, J.~Bohg, and
  A.~Garg.
\newblock Variable impedance control in end-effector space: An action space for
  reinforcement learning in contact-rich tasks.
\newblock In \emph{2019 IEEE/RSJ International Conference on Intelligent Robots
  and Systems (IROS)}, pages 1010--1017. IEEE, 2019.

\bibitem[Ghosh et~al.(2017)Ghosh, Singh, Rajeswaran, Kumar, and
  Levine]{ghosh2017divide}
D.~Ghosh, A.~Singh, A.~Rajeswaran, V.~Kumar, and S.~Levine.
\newblock Divide-and-conquer reinforcement learning.
\newblock \emph{arXiv preprint arXiv:1711.09874}, 2017.

\bibitem[Yu et~al.(2020)Yu, Kumar, Gupta, Levine, Hausman, and
  Finn]{yu2020gradient}
T.~Yu, S.~Kumar, A.~Gupta, S.~Levine, K.~Hausman, and C.~Finn.
\newblock Gradient surgery for multi-task learning.
\newblock \emph{arXiv preprint arXiv:2001.06782}, 2020.

\bibitem[Haarnoja et~al.(2018)Haarnoja, Zhou, Abbeel, and
  Levine]{haarnoja2018soft}
T.~Haarnoja, A.~Zhou, P.~Abbeel, and S.~Levine.
\newblock Soft actor-critic: Off-policy maximum entropy deep reinforcement
  learning with a stochastic actor.
\newblock In \emph{International conference on machine learning}, pages
  1861--1870. PMLR, 2018.

\bibitem[Zhu et~al.(2020)Zhu, Wong, Mandlekar, and
  Mart{\'\i}n-Mart{\'\i}n]{zhu2020robosuite}
Y.~Zhu, J.~Wong, A.~Mandlekar, and R.~Mart{\'\i}n-Mart{\'\i}n.
\newblock robosuite: A modular simulation framework and benchmark for robot
  learning.
\newblock \emph{arXiv preprint arXiv:2009.12293}, 2020.

\bibitem[Todorov et~al.(2012)Todorov, Erez, and Tassa]{todorov2012mujoco}
E.~Todorov, T.~Erez, and Y.~Tassa.
\newblock Mujoco: A physics engine for model-based control.
\newblock In \emph{2012 IEEE/RSJ International Conference on Intelligent Robots
  and Systems}, pages 5026--5033. IEEE, 2012.

\bibitem[Andrychowicz et~al.(2017)Andrychowicz, Wolski, Ray, Schneider, Fong,
  Welinder, McGrew, Tobin, Abbeel, and Zaremba]{andrychowicz2017hindsight}
M.~Andrychowicz, F.~Wolski, A.~Ray, J.~Schneider, R.~Fong, P.~Welinder,
  B.~McGrew, J.~Tobin, P.~Abbeel, and W.~Zaremba.
\newblock Hindsight experience replay.
\newblock \emph{arXiv preprint arXiv:1707.01495}, 2017.

\bibitem[Zhang et~al.(2020)Zhang, Sharma, Liang, and Kroemer]{zhang2020modular}
K.~Zhang, M.~Sharma, J.~Liang, and O.~Kroemer.
\newblock A modular robotic arm control stack for research: Franka-interface
  and frankapy.
\newblock \emph{arXiv preprint arXiv:2011.02398}, 2020.

\end{thebibliography}
